\documentclass{article}

\usepackage{arxiv}

\usepackage[utf8]{inputenc} 
\usepackage[T1]{fontenc}    
\usepackage{hyperref}       
\usepackage{url}            
\usepackage{booktabs}       
\usepackage{amsfonts}       
\usepackage{nicefrac}       
\usepackage{tabularx}
\usepackage{microtype}      
\usepackage{lipsum}
\usepackage{graphicx}
\usepackage{subcaption}
\usepackage{verbatim}
\hypersetup{final}

\newcommand{\xcoord}{\mathbf{x}}
\newcommand{\etal}{\textit{et al.}}
\newcommand{\airlight}{\textit{airlight }}

\title{Reconstruction Loss Minimized FCN for Single Image Dehazing}

\author{
  Shirsendu Sukanta Halder\\
  Department of Computer Science and Engineering\\
  Indian Institute of Technology Roorkee\\
  Roorkee, India 247667 \\
  \texttt{shalder@cs.iitr.ac.in} \\
   \And
 Sanchayan Santra \\
  Electronics and Communication Sciences Unit\\
  Indian Statistical Institute Calcutta\\
  Kolkata, India 700108 \\
  \texttt{sanchayansantra@gmail.com} \\
  \And
 Bhabatosh Chanda \\
  Electronics and Communication Sciences Unit\\
  Indian Statistical Institute Calcutta\\
  Kolkata, India 700108 \\
  \texttt{chanda@isical.ac.in} \\
}

\begin{document}
\maketitle

\begin{abstract}
Haze and fog reduce the visibility of outdoor scenes as a veil like semi-transparent layer appears over the objects. As a result, images captured under such conditions lack contrast. Image dehazing methods try to alleviate this problem by recovering a clear version of the image. In this paper, we propose a Fully Convolutional Neural Network based model to recover the clear scene radiance by estimating the environmental illumination and the scene transmittance jointly from a hazy image. The method uses a relaxed haze imaging model to allow for the situations with non-uniform illumination. We have trained the network by minimizing a custom-defined loss that measures the error of reconstructing the hazy image in three different ways. Additionally, we use a multilevel approach to determine the scene transmittance and the environmental illumination in order to reduces the dependence of the estimate on image scale. Evaluations show that our model performs well compared to the existing \textit{state-of-the-art} methods. It also verifies the potential of our model in diverse situations and various lighting conditions.

\end{abstract}


\section{Introduction}
\begin{figure}
    \centering
         \includegraphics[width=0.29\linewidth]{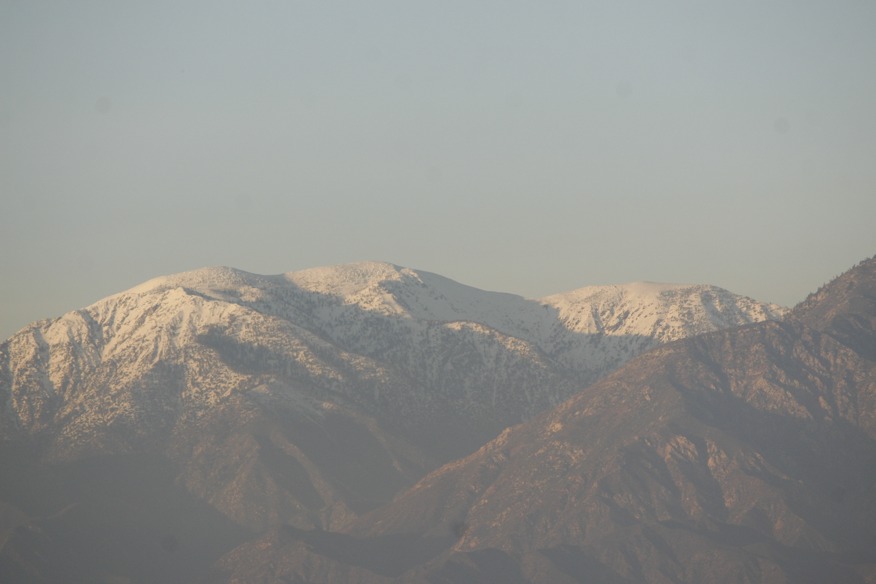}
         \includegraphics[width=0.29\linewidth]{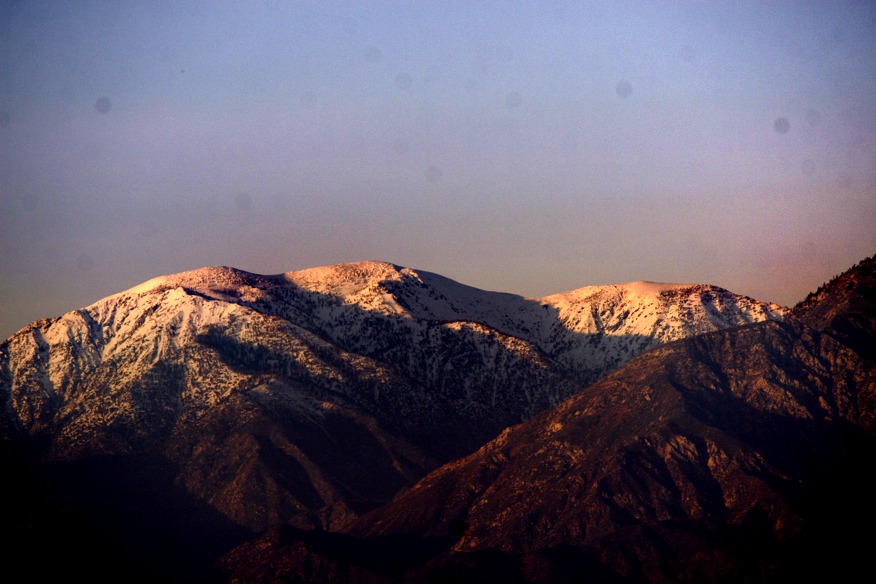}
    \caption{Removal of haze by estimating the scene transmittance and \airlight map using our proposed method}
    \label{fig:firstfigure}
\end{figure}

Images captured in outdoor scenarios are frequently affected by natural phenomena like haze or fog.  The consequences include, degradation of the scene visibility and colour shift of the image. These effects occur due to the presence of minuscule particles in the atmosphere which hamper the passage of light by absorption and reflection \cite{koschi}. In addition to hampering the passage of light, these particles also create a semi-transparent layer of light that affects the visibility. This layer is referred to as \airlight and it directly depends on the transmittance of the medium. The technique of reconstructing clear haze-free images by mitigating the deteriorating effects of haze is known as image dehazing as given in Fig. \ref{fig:firstfigure}.

The problem of image dehazing is an ill-posed one because, the degradation due to haze depends on the scene depth which is non-uniform and unknown at different positions of the image. The primary works on image dehazing took the route of image contrast enhancement \cite{contrast1, oakley}. Subsequently, diverse methods have been proposed in the literature that models the statistical and physical cues to estimate the scene transmittance and the environmental illumination. Single-image dehazing methods \cite{tan, he, fattal2, berman} are getting a good amount of attention over multiple image-based ones \cite{nayar3, nayar2} due to their practical significance. With the recent success of Convolutional Neural Networks (CNN) in Computer Vision \cite{alexnet, vgg, sr}, the problem of image dehazing have also been explored in the light of CNNs \cite{dehazenet,aodnet,ren}. The main advantage of using CNNs is its ability to learn features from a diverse and large set of data without any human intervention. 

Existing dehazing methods mainly focus on the estimation of scene transmittance and do not stress on the correct estimation of environmental illumination. So, in this work we try to estimate both scene transmittance and environmental illumination from image patches. The contributions of our work can be summarized as follows:



\begin{itemize}
    \item Design of a two-way forked Fully Convolutional Network that simultaneously estimates scene transmittance and environmental illumination. 
    \item A novel custom-defined reconstruction loss that conforms to the atmospheric scattering model.
    \item A multilevel approach for inferring scene transmittance and environmental illuminanation to alleviate the problem of varied scale in input image.
\end{itemize}

The rest of the paper is arranged as follows. Section \ref{sec:relatedwork} describes the related works present in literature. The basics of the image formation in the presence of a scattering medium is described in Section \ref{sec:atmosmodel}. Section \ref{sec:proposedmethod} describes our scene illumination and transmittance estimation network, while Section \ref{sec:dehazingmethod} explains our dehazing method. The training data generation and experimental settings are reported in section \ref{sec:datagenandsettings}. In Section \ref{sec:results}, we provide a comparative analysis (both quantitative and qualitative) of our proposed method. Section \ref{sec:conclusion} consists of concluding remarks.

\section{Related Work}
\label{sec:relatedwork}
Image dehazing is considered a challenging problem to solve as the degradation depends on the depth. The first thing that one can observe in a hazy image is its reduced contrast. Hence, earlier methods approached the problem using image enhancement techniques like contrast enhancement \cite{contrast1, oakley}. These methods fail to produce satisfactory results in practical scenarios as they don't take into account the change in haze density with varying depth. Narasimhan and Nayar \cite{nayar3, nayar2} took the help of multiple images taken under different weather conditions to estimate the depth. This depth is then used for dehazing. Single image methods are receiving a lot of attention these days. These methods incorporate the use of additional priors for depth estimation. The method of He \etal{} \cite{he} is based on the observation that for outdoor clear images, in most local patches there are some pixels with very low intensity in at least one of the color channel.  But during haze this value increases due to the added \airlight. This prior information, denoted as the the \emph{dark channel prior} (DCP), is utilized to estimate the scene transmittance. Tang \etal{} \cite{tang} utilized existing hand-crafted features like local max contrast, dark channel, hue disparity and local max saturation in patches to learn a scene transmittance regressor. Fattal \cite{fattal2} based his work on local color line prior. It says that for clear images, colors in a patch form a line in the RGB space, and this line passes through the origin. Under hazy conditions, this line gets shifted by the \airlight depending on the amount of haze. This information is utilized to estimate the transmittance. The work of Berman \etal{} \cite{berman} relies on the assumption that the colors of a natural haze-free image form a few hundred tight clusters in the RGB space. Under haze, these clusters get elongated and form linear structures. These lines, termed \emph{haze-lines}, are employed to estimate the transmittance factors. Although these methods produce good results for certain images, they fail when their assumptions are broken. 




The recent success of Convolutional Neural Networks (CNN) in the domain of Computer Vision \cite{sr,alexnet} has encouraged its use in the problem of Image Dehazing \cite{aodnet,rankingcnn}. CNN based dehazing methods directly regress on transmittance by learning to extract features from the data, instead of relying on hand-crafted features. Dehazenet proposed by Cai \etal{} \cite{dehazenet} works on image patches similar to Tang \etal{} \cite{tang}, but employ a CNN to extract haze relevant features. Ren \etal{} \cite{ren} works with full images to estimate the transmittance using a Multi-Scale CNN. They use two networks: a coarse network to estimate the transmittance map and a fine network for refining the estimated transmittance. Li \etal{} \cite{aodnet} reformulated the atmospheric scattering model \cite{koschi} so that the model contains a single parameter. This unified parameter integrates both scene transmittance and environmental illumination. This parameter is regressed using a CNN named \emph{AOD-Net}.



All the above mentioned methods restrict themselves to daytime scenes where there is a single light source (the sun). Dehazing night-time scenes are more complicated due to the presence of non-uniform illumination. This is normally handled by using a spatially varying atmospheric map. Li \etal{} \cite{li_night} went a bit further and proposed to add a \emph{glow term} to the imaging model, that takes into account the glow effect around the light sources. The method of \cite{santra2016day} is an unified dehazing method that works for both day and night time images using a relaxed atmospheric model. 


\begin{figure*}
    \centering
    \includegraphics[width=\linewidth]{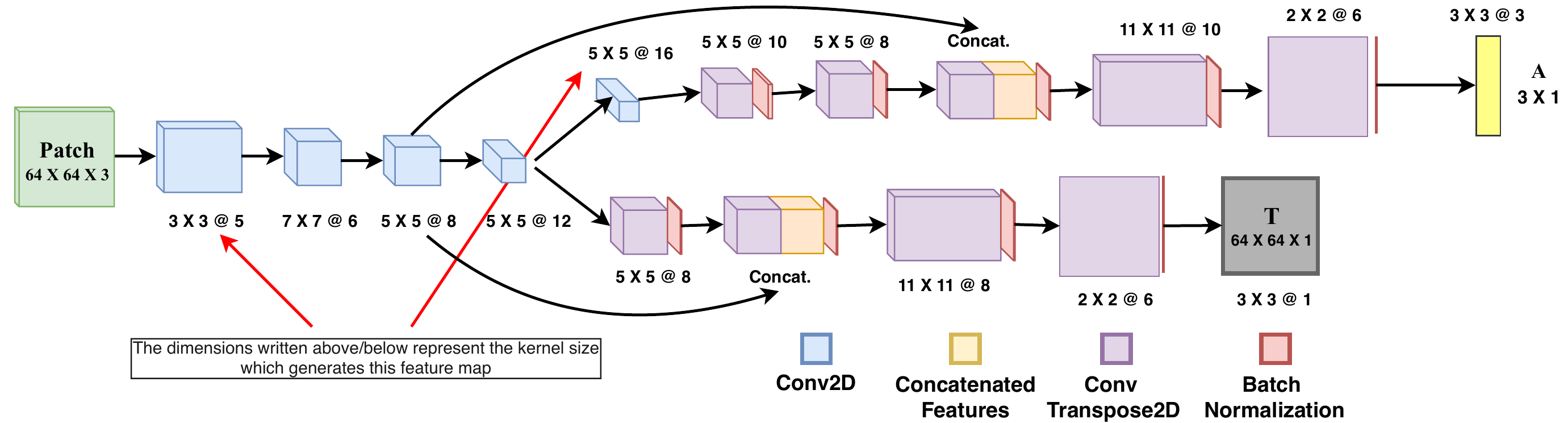}
    \caption{Proposed network for the estimation of the scene transmittance and the environmental illumination}
    \label{fig:model}
\end{figure*}

\section{Imaging Model under Haze}
\label{sec:atmosmodel}
Light that propagates through a medium gets attenuated due to the scattering by the particles present in the medium. The image thus formed has a lesser contrast and a dull colour composition. This phenomenon is modelled by the following equation \cite{koschi}, 
\begin{equation}
    \mathbf{I}(\xcoord) = \mathbf{J}(\xcoord)t(\xcoord) + (1 - t(\xcoord))\mathbf{A},
    \label{eq:haze_image}
\end{equation}

\begin{equation}
    \textnormal{where},~t(\xcoord) = e^{-\beta d(\xcoord)}.
    \label{eq:transmittance}
\end{equation}

Here, $\mathbf{I}(\xcoord)$  is the observed intensity of the image in RGB, whereas $\mathbf{J}(\xcoord)$ is the scene radiance in RGB without the effect of scattering. $\mathbf{A}$ is the global environmental illumination. $t(\xcoord)$ denotes the \emph{scene transmittance} of the light which represents the amount of light that reaches the observer without getting scattered. $\beta$ is the scattering coefficient, and $d(\xcoord)$ is the scene depth. In Eq.~\ref{eq:haze_image}, $\mathbf{A}$ is taken constant assuming the image is taken during the day with overcast sky, which is a common situation for haze and fog. However, that may not always be true. If the sunlight is dominant or if the image is taken during the night, this assumption is violated \cite{nayar3, he}. Therefore, in order to tackle this issue of non-uniform illumination, we use a modified version of Eq.~\ref{eq:haze_image} for our purpose:
\begin{equation}
 \label{eq:opt_model}
 \mathbf{I}(\xcoord) = \mathbf{J}(\xcoord)t(\xcoord) + (1 -t(\xcoord))\mathbf{A}(\xcoord).
\end{equation}
Most of the existing methods work by taking small patches and assume the transmittance to be constant within the patch. The environmental illumination is estimated separately. Our method also works using image patches. However, we aim to estimate the scene transmittance and the environmental illumination simultaneously. But, estimating environmental illumination in a small patch is difficult, as it is hard to differentiate whether the colors are due to the environmental illumination or the object present in the patch. For this reason, we work using bigger patches. But in bigger patches the constant transmittance assumption gets violated. So, for our method we assume transmittance can vary within a patch but the environmental illumination remains constant. This allows for different illumination estimates for different patches. This relaxation helps us to tackle the dehazing of night-time images where we have non-uniform illumination due to artificial lights. Now, to be able to dehaze an image, we need to estimate the constant \airlight within a patch and the scene transmittance map of the same size as the input patch, as we have assumed the transmittance can vary within a patch. This is achieved by using a Fully Convolutional Network (FCN) that estimates both $t(\xcoord)$ and $\mathbf{A}(\xcoord)$ from patches.





\section{Proposed Solution}
\label{sec:proposedmethod}
In the following subsections, we describe the architecture of our proposed network and the loss function we have used to train the network. 

\subsection{Dehazing Network}
\label{sub:network}
For the joint estimation of the environmental illumination and the scene transmittance, we propose a two-way forked FCN (Fig. \ref{fig:model}). The initial four convolution layers of this network extract features for both $t(\xcoord)$ and $\mathbf{A}(\xcoord)$. This is done with the aim of capturing the interdependence between $t(\xcoord)$ and $\mathbf{A}(\xcoord)$. The network then bifurcates into two different sections, one estimates the $t(\xcoord)$ and the other $\mathbf{A}(\xcoord)$. The number of transposed convolutional layers is kept the same as the number of convolutional layers in each path. Some skip connections has also been added in each forked path. These skip connections help to compensate for the loss of fine-scale detail due to convolutions. We use the \emph{tanh} activation function for all the layers, except for the last layer in both the paths. In the last layer we use \emph{sigmoid} to squash the output values between $[0,1]$. Every transposed convolutional layer is followed by a Batch-Normalization layer to reduce over-fitting.



\subsection{Reconstruction Error Minimized Loss}
\label{sub:lossfunction}
To train a regressor, it is common to choose \emph{mean squared error} (MSE) as the loss function. But it is not a good choice for image dehazing. A small error in the estimated $t(\xcoord)$ can have a substantial impact on the dehazed output. This becomes significant as the value of t goes to 0, that is in the areas with dense haze. To address this problem we have formulated a custom loss function based on the imaging model (Eq.~\ref{eq:haze_image}). We define the total loss as follows.
\begin{equation}
    L=\frac{1}{N} \sum_{\xcoord} ( \eta(\xcoord)L_1(\xcoord) + L_2(\xcoord) + \eta(\xcoord) L_3(\xcoord)),
    \label{eq:total_loss}
\end{equation}
where,
\begin{center}
    $L_{1}(\xcoord) = |\mathbf{I}(\xcoord)-\mathbf{J}(\xcoord)t'(\xcoord)-(1-t'(\xcoord))\mathbf{A}_p(\xcoord)|$,\\
    $L_{2}(\xcoord) = |\mathbf{I}(\xcoord)-\mathbf{J}(\xcoord)t_p(\xcoord)-(1-t_p(\xcoord))\mathbf{A'}(\xcoord)|$,\\
    $L_{3}(\xcoord) = |\mathbf{I}(\xcoord)-\mathbf{J}(\xcoord)t_p(\xcoord)-(1-t_p(\xcoord))\mathbf{A}_p(\xcoord)|$,
\end{center}

\begin{equation}
    \textnormal{and}~\eta(\xcoord) = \Big(1 - \frac{ e^{\gamma t'(\xcoord)}-1}{e^{\gamma}-1}\Big) ;\gamma \geq 1.
    \label{eq:eta_gamma}
\end{equation}
Here $\mathbf{I}(\xcoord)$, $\mathbf{J}(\xcoord)$ are the input hazy image and ground truth clean image respectively. $t'(\xcoord)$ and $\mathbf{A'}(\xcoord)$ are ground truth transmittance map and ground truth environmental illumination, while $t_p(\xcoord)$ and $\mathbf{A}_p(\xcoord)$ are the transmittance map and  environmental illumination obtained from the network. $N$ denotes the number of pixel in the patch. With these three losses we try to minimize the loss of reconstructing hazy image from the ground truth haze-free images in three different ways.
\begin{enumerate}
    \item [($L_1$)] : with predicted $\mathbf{A}(\xcoord)$, but ground-truth $t(\xcoord)$
    \item [($L_2$)] : with predicted $t(\xcoord)$, but ground-truth $\mathbf{A}(\xcoord)$
    \item [($L_3$)] : with predicted $t(\xcoord)$ and $\mathbf{A}(\xcoord)$
\end{enumerate}
Using only $L_3$ the network is likely to get stuck at trivial solutions like $t(\xcoord) = 0$ and $\mathbf{A}(\xcoord) = \mathbf{I}(\xcoord)$. The addition of $L_1$ and $L_2$ prevents these situations. These two also guide the prediction towards the actual values. But, on the other hand as the value of $t(\xcoord)$ increases, the effect of $\mathbf{A}(\xcoord)$ in the image reduces, because $(1-t(\xcoord))\mathbf{A}(\xcoord)$ goes towards zero. In this case the network can learn to output some arbitrary $\mathbf{A}(\xcoord)$. 
To address this issue we have reduced the importance of $L_1$ and $L_3$ using the term $\eta(\xcoord)$ as we are using the predicted $\mathbf{A(\xcoord)}$ in these cases. We don't do this for $L_2$ as we are using ground-truth $\mathbf{A}(\xcoord)$ in it. Note that, for $0 \le t(\xcoord) \le 1$ and $\gamma \geq 1$, it can be easily shown that $0 \leq \eta \leq 1$. For our method we have taken $\gamma$ to be 15.




\section{Dehazing Method}
\label{sec:dehazingmethod}
Our dehazing method consists of four main steps:
\begin{enumerate}
    \item Multilevel estimation of $t(\xcoord)$ and $\mathbf{A}(\xcoord)$.
    \item $t(\xcoord)$ and $\mathbf{A}(\xcoord)$ aggregation.
    \item Regularization and interpolation.
    \item Haze-free image recovery.
\end{enumerate}
Each step is described in detail in the following subsections.

\subsection{Multilevel Estimation of \texorpdfstring{$t(\xcoord)$}{Lg} and \texorpdfstring{$\mathbf{A}(\xcoord)$}{Lg}}

We first estimate $t(\xcoord)$ and $A(\xcoord)$ from different patches of the input image using our dehazing network. For that we consider only patches (overlapping) that are not smooth. Now, depending on the resolution of the image, a same sized patch can cover a different amount of area. So, for a given input image, if we just take patches out of it and feed to the network, the accuracy of the obtained estimate can vary depending on the resolution of the input image. For this reason, we estimate the two parameters ($t(\xcoord)$ and $\mathbf{A}(\xcoord)$) at multiple levels by taking patches of different sizes from the input image. 

For the multilevel estimation, we begin with a patch size of $P\times P$. Where, $P = \min(H, W)$ for an image of size $H \times W$. These $P\times P$ patches are resized to $\omega \times \omega$ and fed to the dehazing network. This resize is being done as the network takes $\omega \times \omega$ patches as input. The obtained $t(\mathbf{x})$ and $A(\mathbf{x})$ for a patch is re-sized back to the original size ($P \times P$). Now, for the second level we take patches of size $\frac{P}{2}\times\frac{P}{2}$ and estimate the two parameters in the same way. This procedure is repeated until the size of the patch we are taking falls below $\omega \times \omega$. The number of levels thus obtained is given by:
\begin{equation}
    \label{eq:levels}
    m = \lfloor (\log_2(\min(H,W)) - \log_2(\omega)) + 1 \rfloor
\end{equation}


\subsection{\texorpdfstring{$t(\xcoord)$}{Lg} and \texorpdfstring{$\mathbf{A}(\xcoord)$}{Lg} Aggregation}
The estimates obtained from patches needs to be aggregated to obtain the full sized $t(\xcoord)$ and $\mathbf{A}(\xcoord)$ maps before we are able to dehaze an image. Therefore, in each level we first aggregate the patches by averaging the values to obtain full sized maps. Now, the estimates obtained at each level are aggregated using weighted average to get the overall estimates as follows:
\begin{equation}
    \label{eq:t_estimation}
    t(\xcoord) = \frac{\sum_{i=1}^{m} w_i^t t_i(\xcoord)}{\sum_{i=1}^{m}w_i^t}
\end{equation}
\begin{equation}
    \label{eq:A_estimation}
    \mathbf{A}(\xcoord) = \frac{\sum_{i=1}^{m} w_i^A \mathbf{A}_i(\xcoord)}{\sum_{i=1}^{m}w_i^A}
\end{equation}
Here, $t_i(\xcoord)$ and $\mathbf{A}_i(\xcoord)$ represent the maps obtained at $i^{th}$ level,  while $m$ denotes the total number of levels. Note that, for our case we have taken the weights as 1. 

\subsection{Regularization and Interpolation}
\begin{figure}
    \centering
    \begin{subfigure}[t]{0.35\linewidth}
        \includegraphics[width=\linewidth]{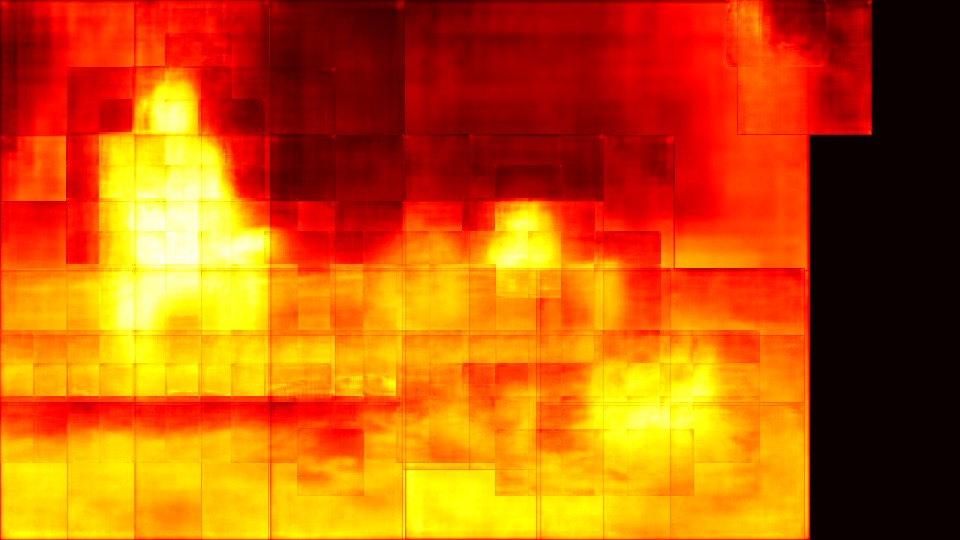}\\
        \includegraphics[width=\linewidth]{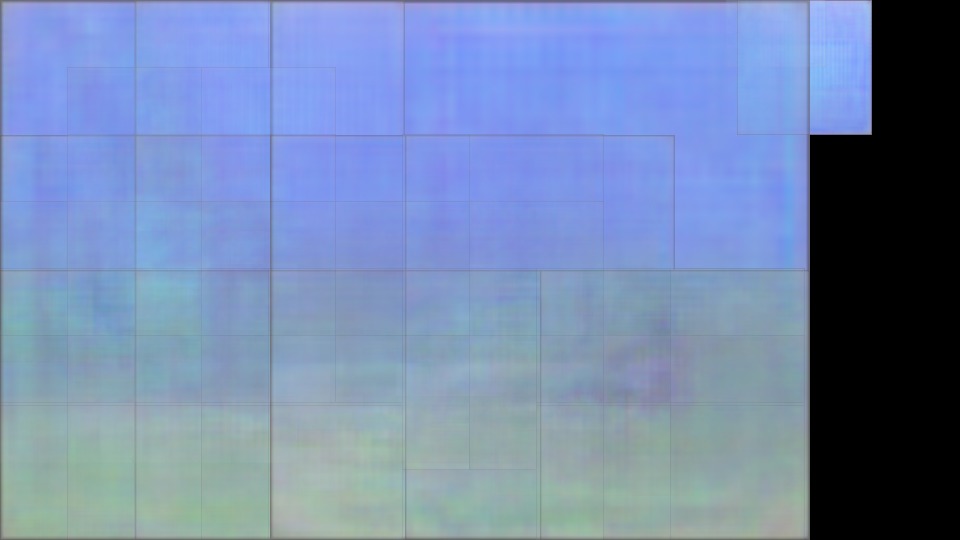}
        \caption{Without regularization}
    \end{subfigure}
    \begin{subfigure}[t]{0.35\linewidth}
        \includegraphics[width=\linewidth]{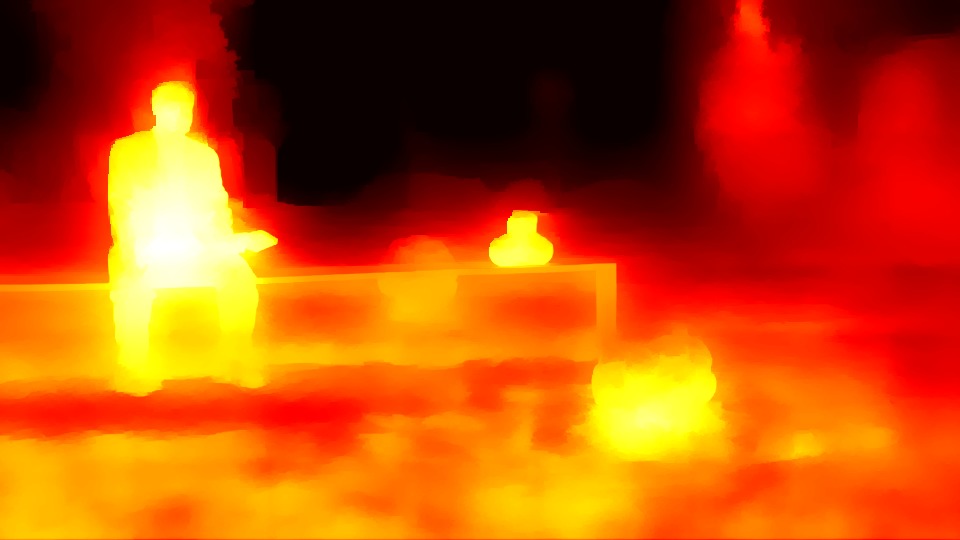}\\
         \includegraphics[width=\linewidth]{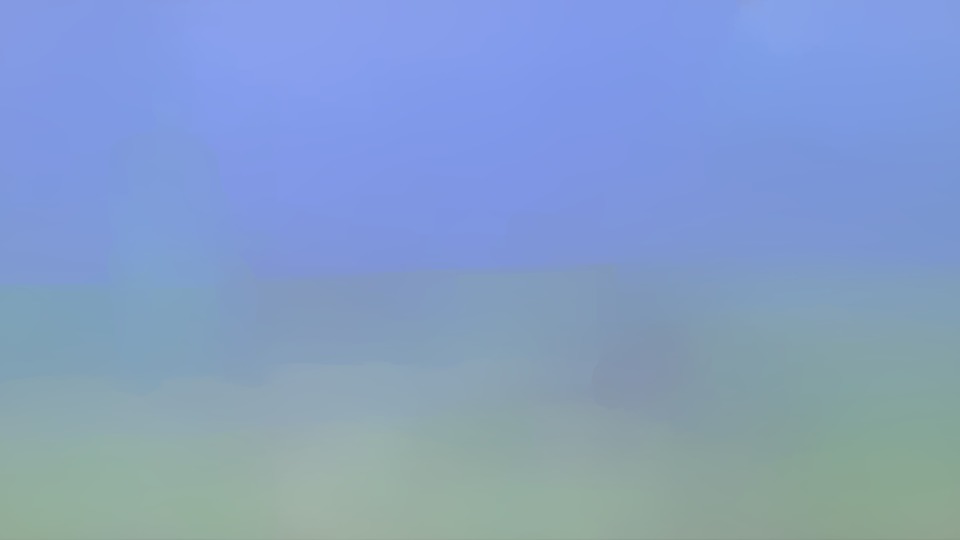}
         \caption{With regularization}
    \end{subfigure}
    \caption{Effect of using regularizer on the $t(\xcoord)$ and $\mathbf{A}(\xcoord)$ map}
    \label{fig:regularizer_effect}
\end{figure}

Due to the patch-based processing of the images, the overall estimated $t(\xcoord)$ and $\mathbf{A}(\xcoord)$ contain halos at patch borders (Fig. \ref{fig:regularizer_effect}). Also, we have not considered smooth patches in the estimation step. So, at those pixels we don't have the estimates. To alleviate this problem, we interpolate and regularize the overall estimates. This is done using a Laplacian based regularizer similar to Fattal \cite{fattal2}. This is achieved by maximizing the following Gauss-Markov random field model,
\begin{equation}
    P(a)\propto\exp \Big(-\sum_\xcoord s(\xcoord) (a(\xcoord)-\hat{a}(\xcoord))^2-\sum_\xcoord\sum_{\mathbf{y}\in N_\xcoord} \frac{(a(\xcoord)-a(\mathbf{y})^2)}{||I(\xcoord)-I(\mathbf{y})||^2}\Big),
\end{equation}
where $\hat{a}(\xcoord)$ the overall estimate and $N_\xcoord$ denotes the neighborhood of $\xcoord$. $s(\xcoord)$ is 1 where estimates are available and 0 otherwise. This specific regularizer helps in smoothing the estimates at the same time retaining sharp profile along the edges depending on the input image. Both the transmittance and environmental illumination (each channel separately) is smoothed with this model.


\subsection{Haze-free Image Recovery}
Using the smoothed $t$ and $\mathbf{A}$ map,we recover the dehazed image using the following equation, 
\begin{equation}
 \label{eq:dehazing_eq}
 \mathbf{J}(\xcoord) = \frac{\mathbf{I}(\xcoord) - (1 - t(\xcoord)) \mathbf{A}(\xcoord)}{\max\{0.1, t(\xcoord)\}}
\end{equation}
Note that, to ensure the value of $\mathbf{J}(\xcoord)$ stays within the valid range, we clip the values of the $t(\xcoord)$ present in the denominator. 

\begin{table*}
\centering
\caption{Quantitative comparison of PSNR/SSIM/CIEDE2000 values on images from Fattal's dataset}
\label{tb:fattalquantitative}
\begin{tabular}{cccccc}
\hline
& Dehazenet \cite{dehazenet} & Berman \etal \cite{berman} & AOD-Net \cite{aodnet} & MSCNN \cite{ren} & Ours\\ \hline
Church & 14.64/0.82/20.45 & 15.69/0.88/16.91 & 9.44/0.61/34.64 & 14.18/0.85/20.26 & \textit{\textbf{18.52}}/\textit{\textbf{0.89}}/\textit{\textbf{13.54}}\\
Couch & 16.71/0.83/14.34 & 17.28/0.86/14.19 & 16.79/0.82/17.33 & 18.02/\textbf{0.87}/12.92 & \textit{\textbf{18.69}}/\textit{\textbf{0.87}}/\textit{\textbf{12.31}}\\
Flower1 & \textbf{19.82}/\textbf{0.94}/16.72 & 12.15/0.71/21.00 & 12.21/0.79/29.42 & 9.08/0.42/24.65 & \textit{15.95}/\textit{0.72}/\textit{15.45}\\
Flower2 & 19.44/\textbf{0.91}/15.37 & 11.86/0.67/21.17 & 13.13/0.78/25.27 & 10.82/0.59/22.46 & \textit{\textbf{20.58}}/\textit{0.88}/\textit{\textbf{11.74}}\\
Lawn1 & 13.80/0.81/23.01 & 14.78/\textbf{0.83}/\textbf{17.93} & 11.33/0.67/31.74 & 14.38/0.80/21.00 & \textit{\textbf{16.05}}/\textit{\textbf{0.83}}/\textit{18.65}\\
Lawn2 & 13.61/0.81/22.47 & 15.32/\textbf{0.85}/\textbf{17.81} & 10.98/0.66/31.70 & 13.30/0.76/22.27 & \textit{\textbf{16.55}}/\textit{\textbf{0.85}}/\textit{19.78}\\
Mansion & 17.39/0.84/17.42& 17.34/0.87/15.84 & 14.23/0.69/24.01 & 17.70/0.87/17.53 & \textit{\textbf{20.71}}/\textit{\textbf{0.93}}/\textit{\textbf{12.08}}\\
Moebius & \textbf{19.18}/\textbf{0.94}/\textbf{16.38} & 14.59/0.83/22.40 & 13.22/0.76/27.61 & 16.38/0.89/19.86 & \textit{16.94}/\textit{0.79}/\textit{16.53}\\
Raindeer & 17.87/0.85/13.73 & 16.60/0.80/15.28 & 16.54/0.79/18.50 & 16.83/0.80/15.49 & \textit{\textbf{20.15}}/\textit{\textbf{0.89}}/\textit{\textbf{13.16}}\\
Road1 & 13.74/0.79/22.20 & 16.33/0.87/19.06 & 11.75/0.65/29.32 & 14.13/0.82/22.22 & \textit{\textbf{17.67}}/\textit{\textbf{0.89}}/\textit{\textbf{18.38}}
\\
Road2 & 13.22/0.77/23.43 & \textbf{18.23}/\textbf{0.89}/16.83 & 11.95/0.61/30.96 & 16.45/0.86/20.18 & \textit{17.49}/\textit{0.78}/\textit{\textbf{16.63}}\\
\hline 
 \textbf{Average}& 16.31/\textbf{0.84}/18.68 & 15.47/0.82/18.03 & 12.87/0.71/27.31 & 14.66/0.77/19.89 & \textit{\textbf{18.18}}/ \textit{\textbf{0.84}}/\textit{\textbf{15.79}} \\\hline
\end{tabular}
\end{table*}

\begin{figure*}
    \centering
    \begin{subfigure}[t]{0.13\linewidth}
        \includegraphics[width=\linewidth]{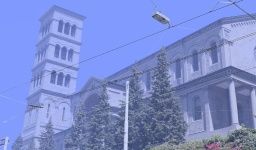}\\
        \includegraphics[width=\linewidth]{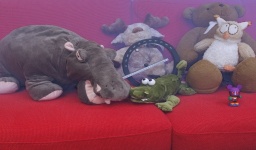}\\
        \includegraphics[width=\linewidth]{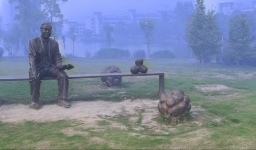} \\
        \includegraphics[width=\linewidth]{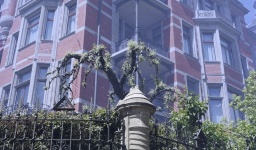} \\
        \includegraphics[width=\linewidth]{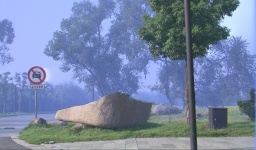}
        \caption{Hazy}
    \end{subfigure}
    \begin{subfigure}[t]{0.13\linewidth}
        \includegraphics[width=\linewidth]{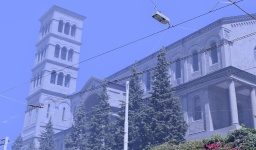}\\
        \includegraphics[width=\linewidth]{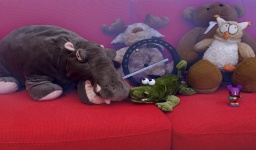}\\
        \includegraphics[width=\linewidth]{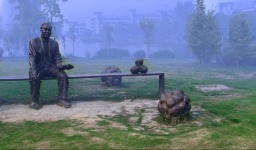} \\ 
        \includegraphics[width=\linewidth]{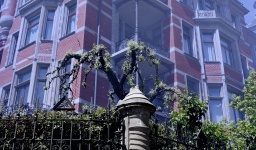} \\ 
        \includegraphics[width=\linewidth]{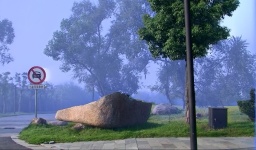}
        \caption{Dehazenet}
    \end{subfigure}
    \begin{subfigure}[t]{0.13\linewidth}
        \includegraphics[width=\linewidth]{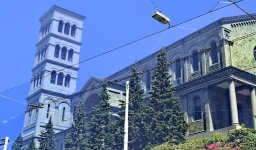}\\
        \includegraphics[width=\linewidth]{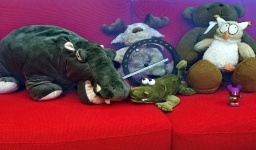}\\
        \includegraphics[width=\linewidth]{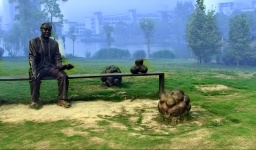} \\ 
        \includegraphics[width=\linewidth]{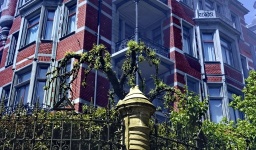} \\
        \includegraphics[width=\linewidth]{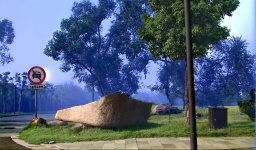}
        \caption{Berman \etal{}}
    \end{subfigure}
    \begin{subfigure}[t]{0.13\linewidth}
        \includegraphics[width=\linewidth]{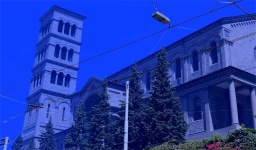}\\
        \includegraphics[width=\linewidth]{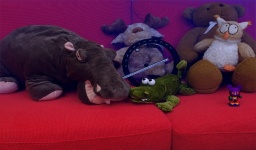}\\
        \includegraphics[width=\linewidth]{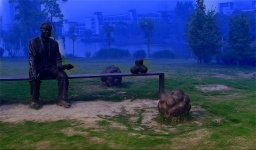} \\ 
        \includegraphics[width=\linewidth]{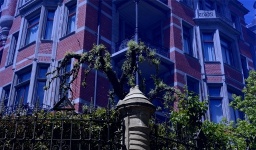} \\ 
        \includegraphics[width=\linewidth]{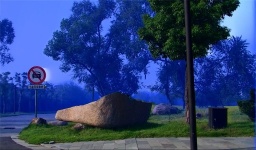}
        \caption{AOD-Net}
    \end{subfigure}
    \begin{subfigure}[t]{0.13\linewidth}
        \includegraphics[width=\linewidth]{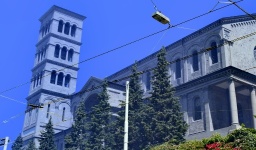}\\
        \includegraphics[width=\linewidth]{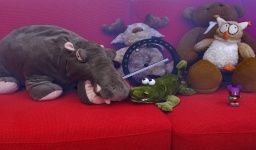}\\
        \includegraphics[width=\linewidth]{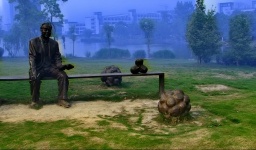} \\ 
        \includegraphics[width=\linewidth]{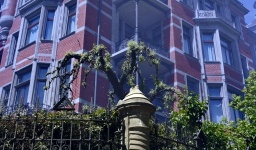} \\ 
        \includegraphics[width=\linewidth]{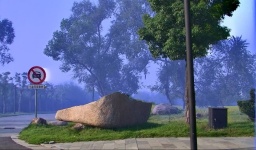}
        \caption{MSCNN}
    \end{subfigure}
    \begin{subfigure}[t]{0.13\linewidth}
        \includegraphics[width=\linewidth]{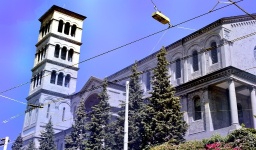}\\
        \includegraphics[width=\linewidth]{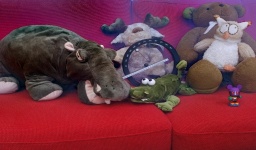}\\
        \includegraphics[width=\linewidth]{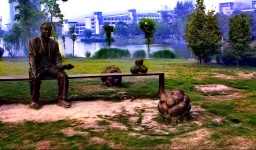} \\ 
        \includegraphics[width=\linewidth]{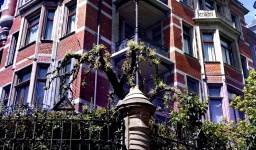} \\ 
        \includegraphics[width=\linewidth]{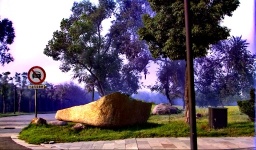}
        \caption{Ours}
    \end{subfigure}
    \begin{subfigure}[t]{0.13\linewidth}
        \includegraphics[width=\linewidth]{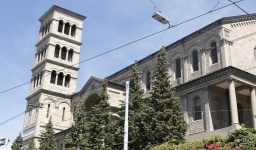}\\
        \includegraphics[width=\linewidth]{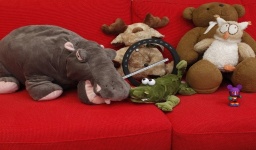} \\
        \includegraphics[width=\linewidth]{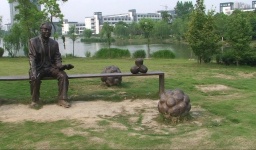} \\
        \includegraphics[width=\linewidth]{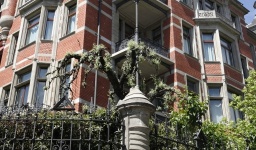}\\
        \includegraphics[width=\linewidth]{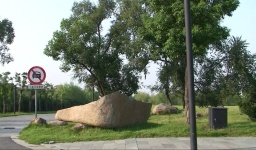}
        \caption{Ground truth}
    \end{subfigure}
    \caption{Visual comparison of images from Fattal's dataset \cite{fattal2}: \emph{Church, Couch, Lawn1, Mansion, Road1}}
    \label{fig:fattalvisual}
\end{figure*}

\section{Data Generation and Settings}
\label{sec:datagenandsettings}
In this section we describe the procedure of the training data generation for our proposed network and the experimental settings under which we evaluate our results. 

\subsection{Data Generation}
\label{sub:datagen}
One of the primary hurdles that one faces while working on image dehazing is the absence of a relevant dataset. It is logistically difficult to acquire a pair of clear and hazy image of the same scene. To circumvent this issue, we have synthesized images with known depth maps to create our training dataset. We utilize Eq. (\ref{eq:transmittance}) to obtain the scene transmittance maps using known depths and Eq. (\ref{eq:opt_model}) to generate the hazy images using the scene transmittance and environmental illumination. We have used the NYU Depth Dataset V2 \cite{nyuv2} for this purpose. NYU-V2 contains 1449 indoor images along with their depth maps, captured using Microsoft Kinect. To generate the the transmittance maps using Eq.~\ref{eq:transmittance}, we have taken $\beta$ in the interval $[0.5,1]$. The usage of $\beta$ beyond this range is avoided, because it results in either very thin or very thick haze. For the environmental illumination, we have taken values between $[0.45, 1]$ for each channel. From the generated hazy images, we have extracted patches and have taken only the ones with variance greater than $0.08$. This is being done on the ground that smooth patches do not contain much information. 



\subsection{Experimental Settings}
\label{sub:experimental}
All the results we report here is obtained on a machine with an Intel\textregistered{} Xeon\textregistered{} 3.1GHz octa core CPU having 64 GB RAM, Nvidia\textregistered{} Tesla\texttrademark{} C2075 and running on Ubuntu 16.04. The dehazing network is built using Keras: The Python Deep Learning library, with Tensorflow as the backend. We have used the Adagrad \cite{duchi2011adaptive} optimizer with a learning rate of $0.01$. Under these settings, the network has been trained with $64\times 64$ sized input patches for 150 epochs with batch size of 32.


\section{Results and Evaluations}
\label{sec:results}
In this section we have evaluated the effectiveness of our method on quantitative as well as qualitative grounds. We have reported ours results on synthetic images, real-world images and also results on night-time hazy images. We have compared the results with existing state-of-the-art methods like Dehazenet \cite{dehazenet}, Berman \etal \cite{berman}, AOD-Net \etal \cite{aodnet} and MSCNN \cite{ren}. Apart from Berman \etal \cite{berman}, rest are CNN-based methods. The results we report here are generated using the code given by the respective authors.

\subsection{Synthetic Images}
\label{sub:synthetic}
In order to evaluate our proposed method in qualitative and quantitative terms, we use two different dataset with synthetically generated hazy images:  Fattal's dataset \cite{fattal2} and Middlebury part of the D-hazy dataset \cite{dhazy}. We don't use its NYU section as the network has been trained with it. Fattal's dataset \cite{fattal2} consists of synthetic indoor and outdoor hazy images and their corresponding haze-free images. Middlebury part of D-hazy dataset contains high resolution indoor images. We provide evaluations of some images from both the datasets (\cite{fattal2} \& \cite{middlebury}). 



\begin{table*}
\centering
\caption{Quantitative comparison of PSNR/SSIM/CIEDE2000 values on images from Middlebury dataset}
\label{tb:middquantitative}
\begin{tabular}{cccccc}
\hline
\multicolumn{1}{c}{} & \multicolumn{1}{c}{Dehazenet \cite{dehazenet}} & \multicolumn{1}{c}{Berman \etal\cite{berman}} & \multicolumn{1}{c}{AOD-Net \cite{aodnet}} & \multicolumn{1}{c}{MSCNN \cite{ren}} & \multicolumn{1}{c}{Ours}\\ \hline
Cable & 8.14/\textbf{0.64}/29.46 & \textbf{9.94}/0.60/\textbf{24.11} & 6.95/0.6/32.64 & 7.65/0.62/29.44 & 7.88/0.6/32.13\\
Couch & 11.49/0.62/19.01 & \textbf{13.77}/\textbf{0.68}/\textbf{16.50} & 10.56/0.61/21.11 & 10.13/0.60/23.16 & 12.13/0.66/19.97\\
Piano & 15.75/0.78/15.62 & 15.07/0.76/15.146 & 13.89/0.74/\textbf{13.93} & 12.39/0.70/17.34 & \textbf{15.79}/\textbf{0.75}/15.07\\
Playroom & 14.57/0.78/15.17 & \textbf{17.64}/\textbf{0.80}/\textbf{10.10} & 13.24/0.76/14.25 & 13.42/0.76/15.07 & 14.52/0.6/14.85\\
Shopvac & 8.00/0.64/30.70 & \textbf{11.58}/\textbf{0.75}/\textbf{19.25} & 6.89/0.61/35.22 & 7.62/0.60/32.43 & 8.89/0.568/26.59\\ \hline
 \textbf{Average}& 11.59/0.69/21.99 & \textbf{13.60}/\textbf{0.72}/\textbf{17.02} & 10.31/0.67/23.43 & 10.24/0.60/23.49 & 11.81/0.67/21.75\\\hline
\end{tabular}
\end{table*}
\begin{figure*}
    \centering
    \begin{subfigure}[t]{0.13\linewidth}
        \includegraphics[height=0.6in, width=\linewidth]{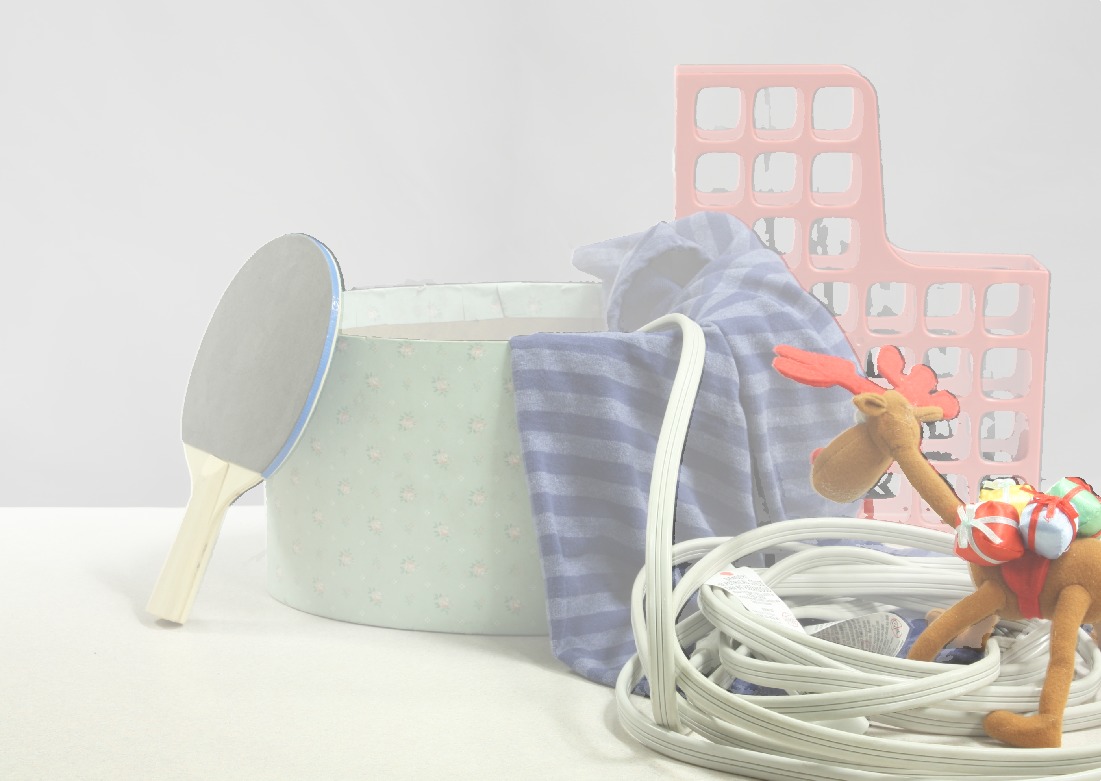} \\
        \includegraphics[height=0.6in, width=\linewidth]{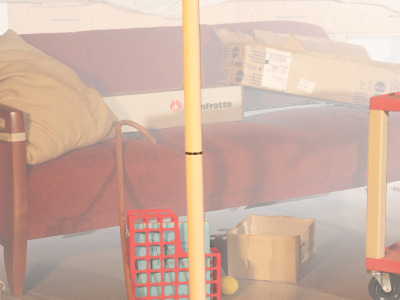} \\
        \includegraphics[height=0.6in, width=\linewidth]{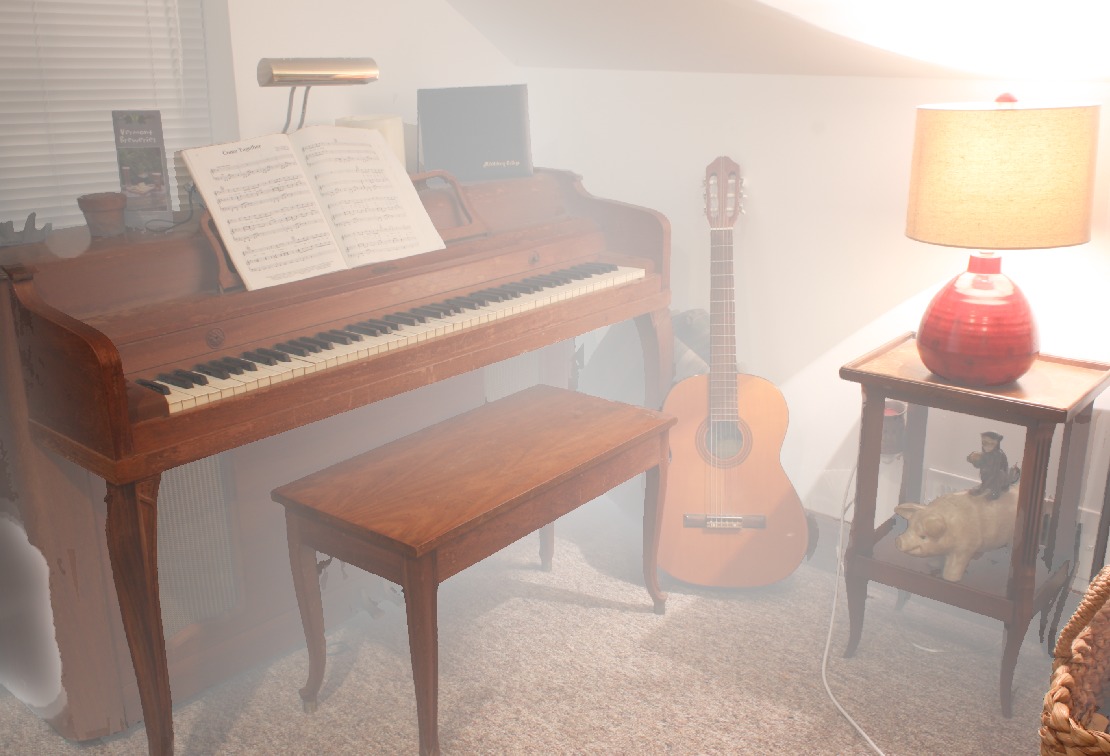}\\
        \includegraphics[height=0.6in, width=\linewidth]{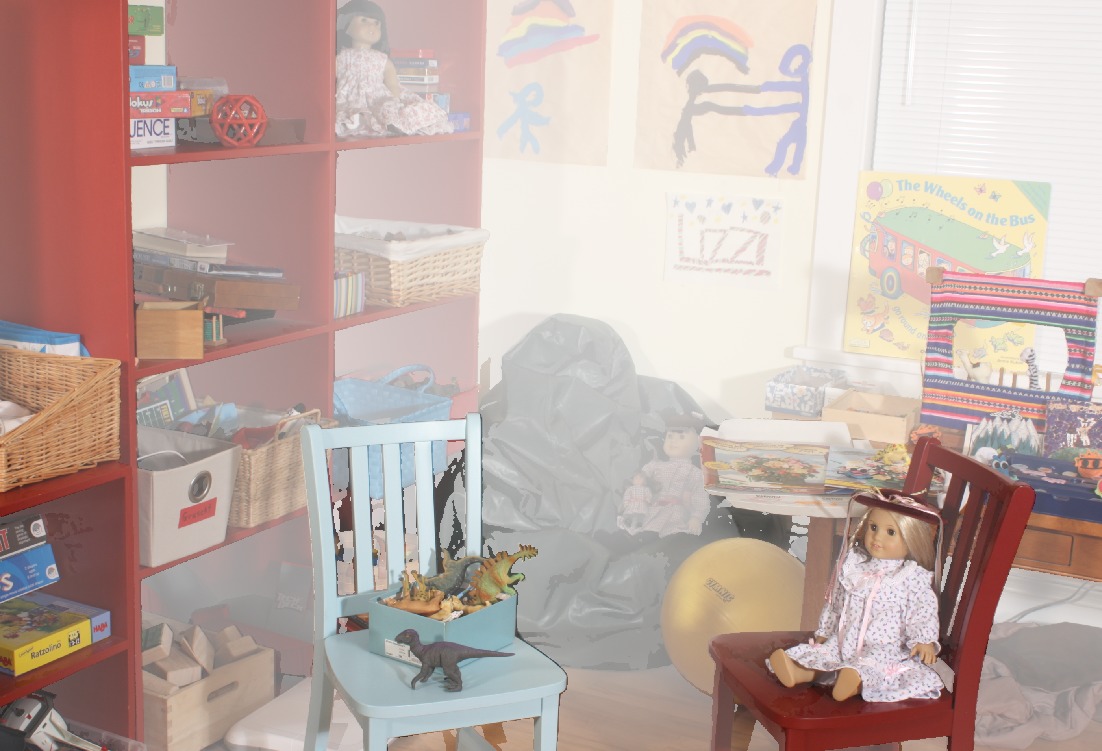} \\
        \includegraphics[height=0.6in, width=\linewidth]{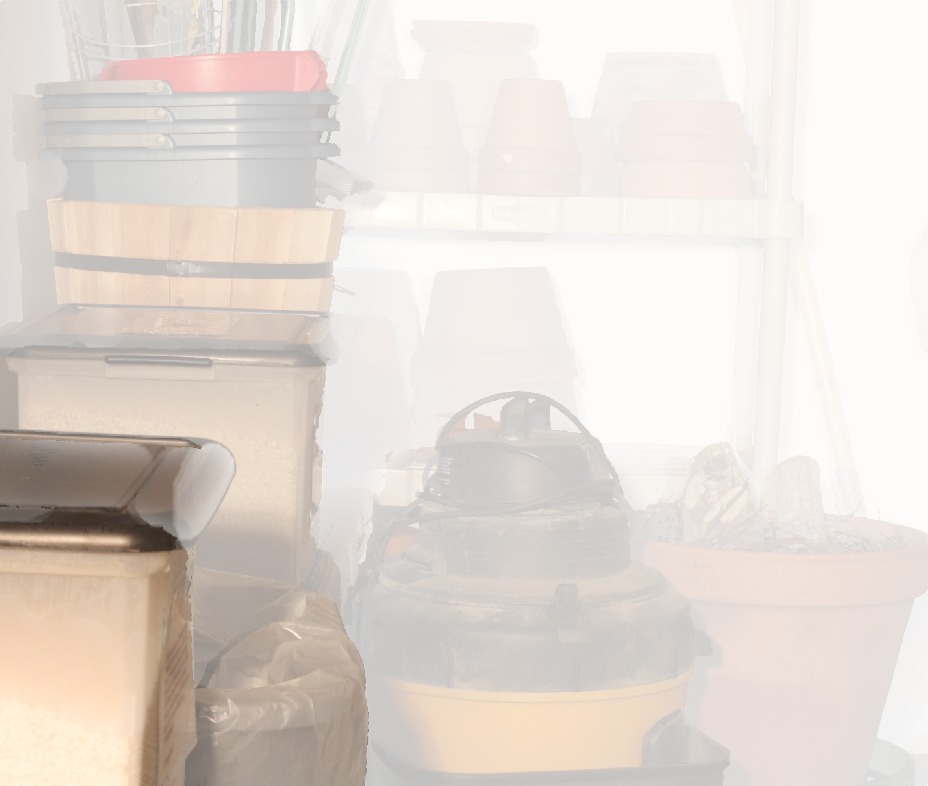} 
        \caption{Hazy}
    \end{subfigure}
    \begin{subfigure}[t]{0.13\linewidth}
        \includegraphics[height=0.6in,width=\linewidth]{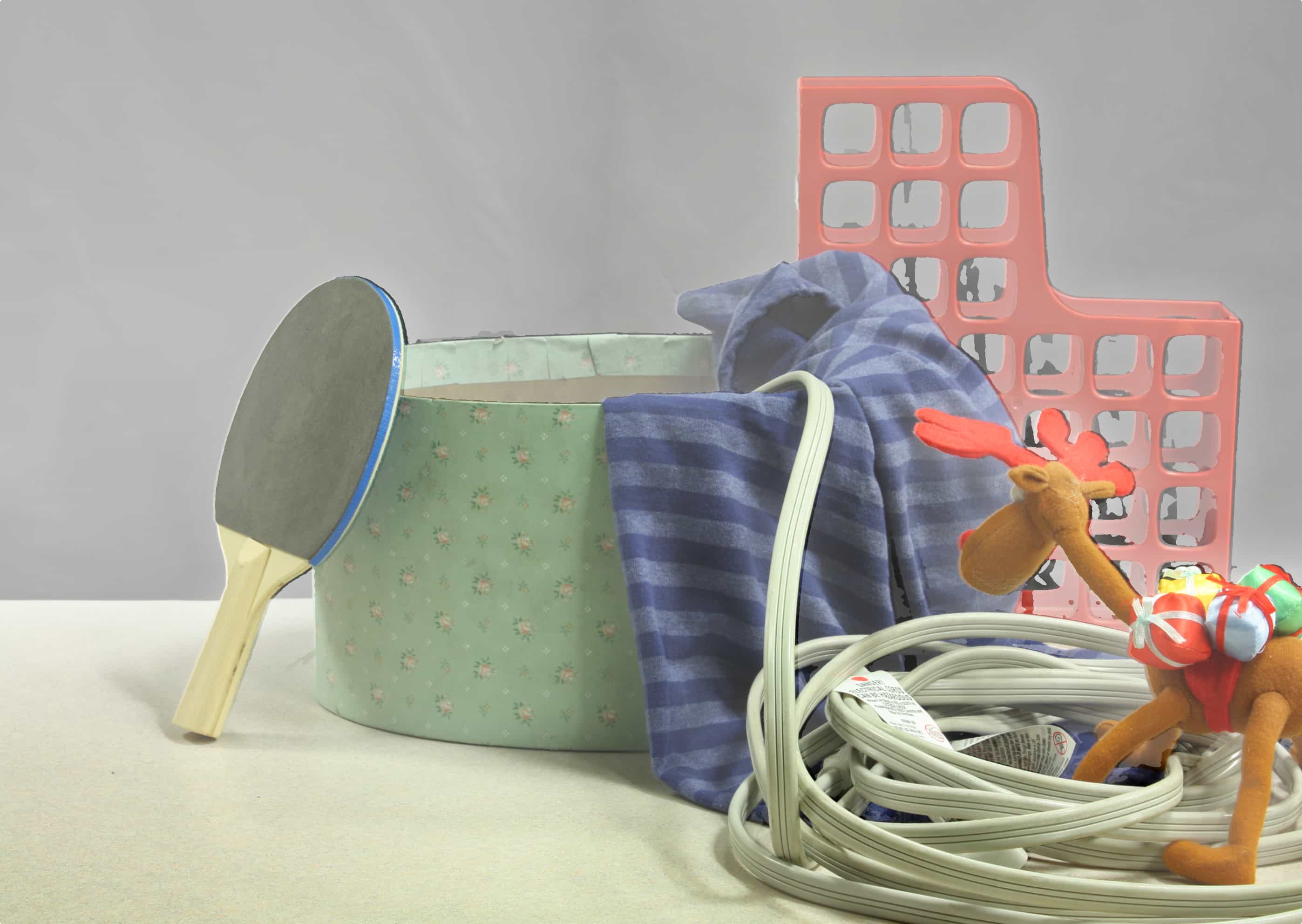} \\
        \includegraphics[height=0.6in,width=\linewidth]{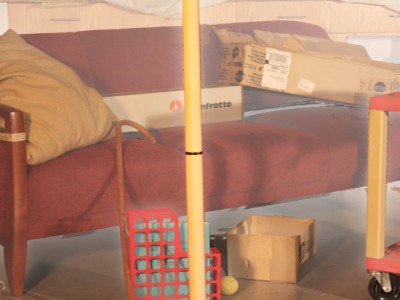} \\
        \includegraphics[height=0.6in,width=\linewidth]{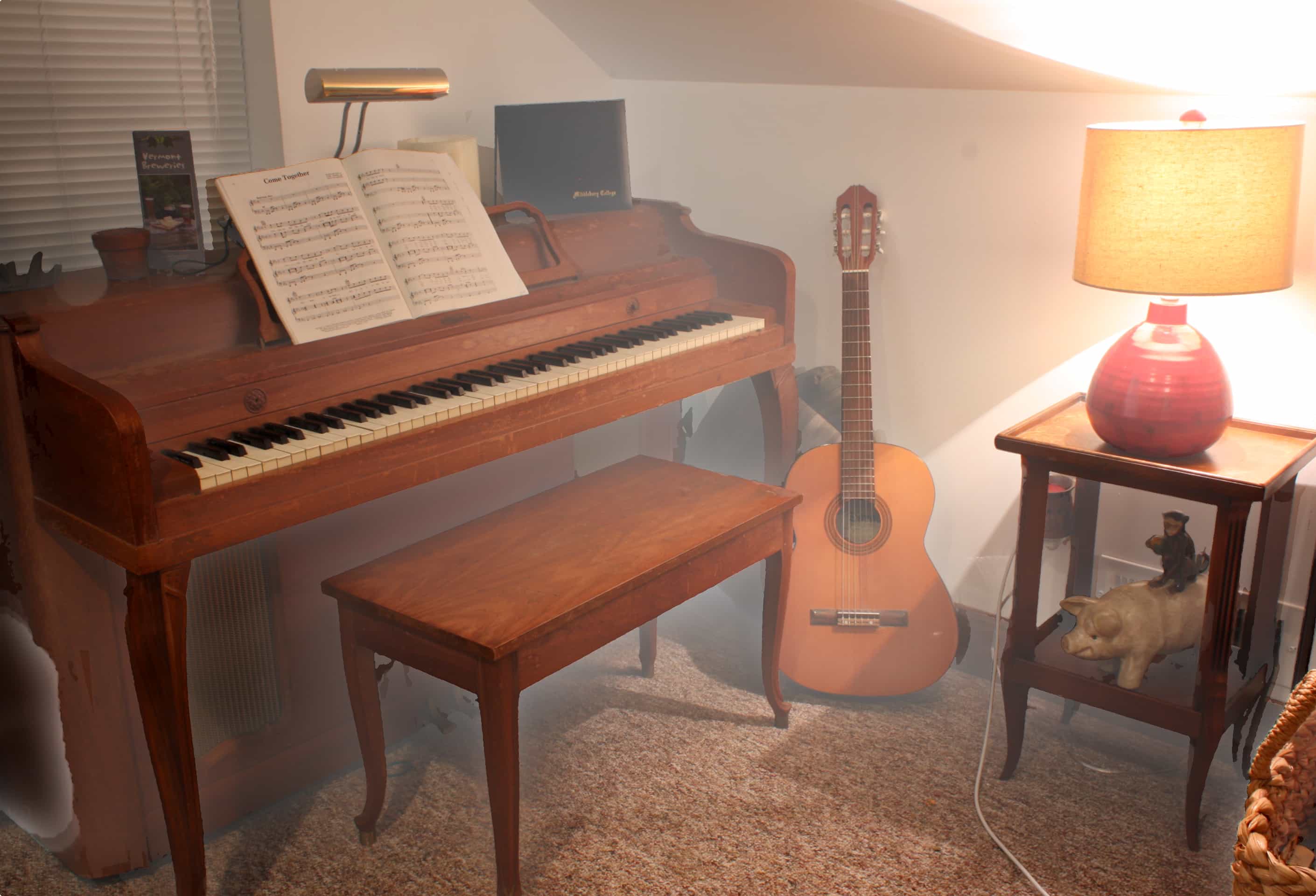}\\
        \includegraphics[height=0.6in,width=\linewidth]{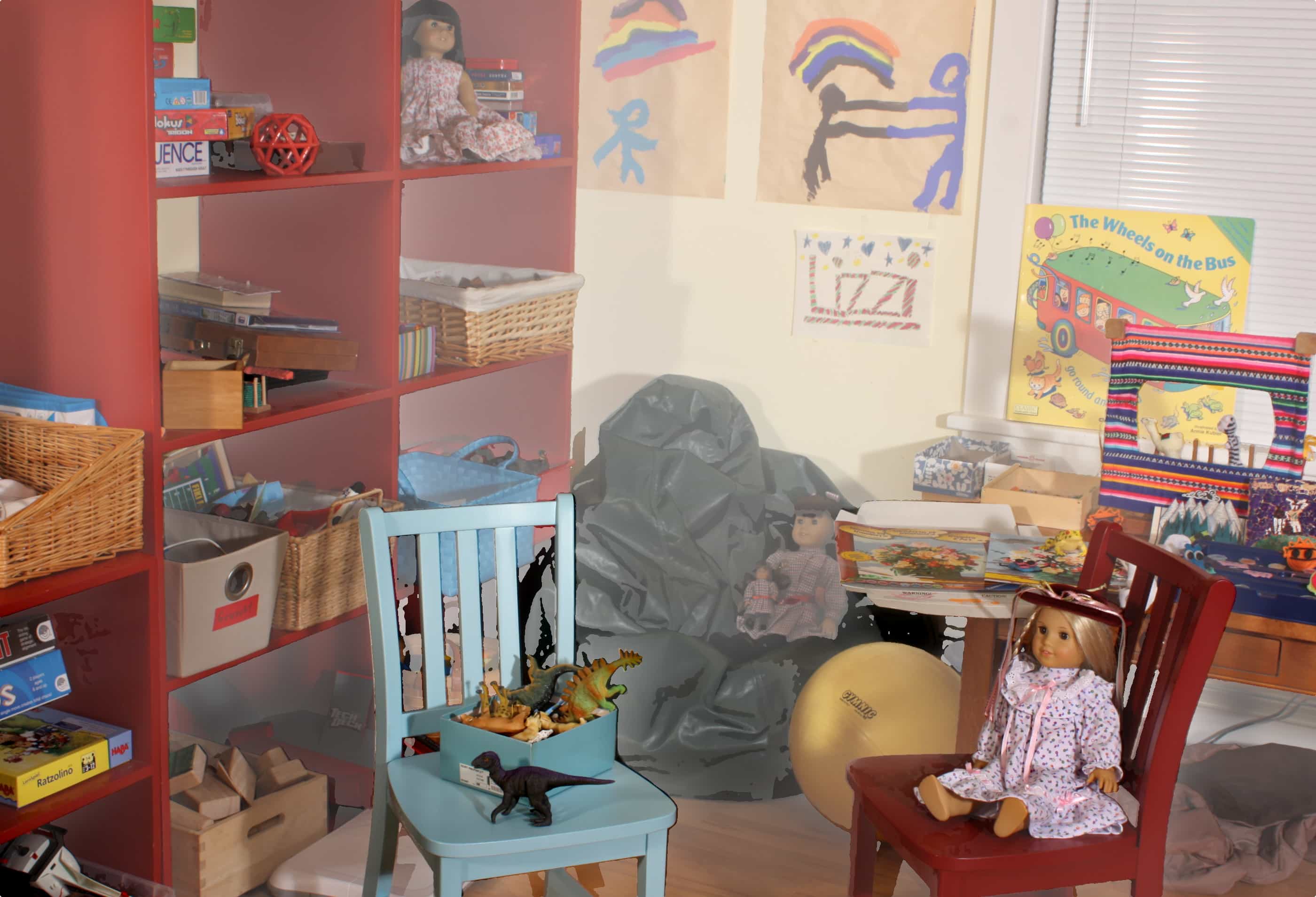} \\
        \includegraphics[height=0.6in,width=\linewidth]{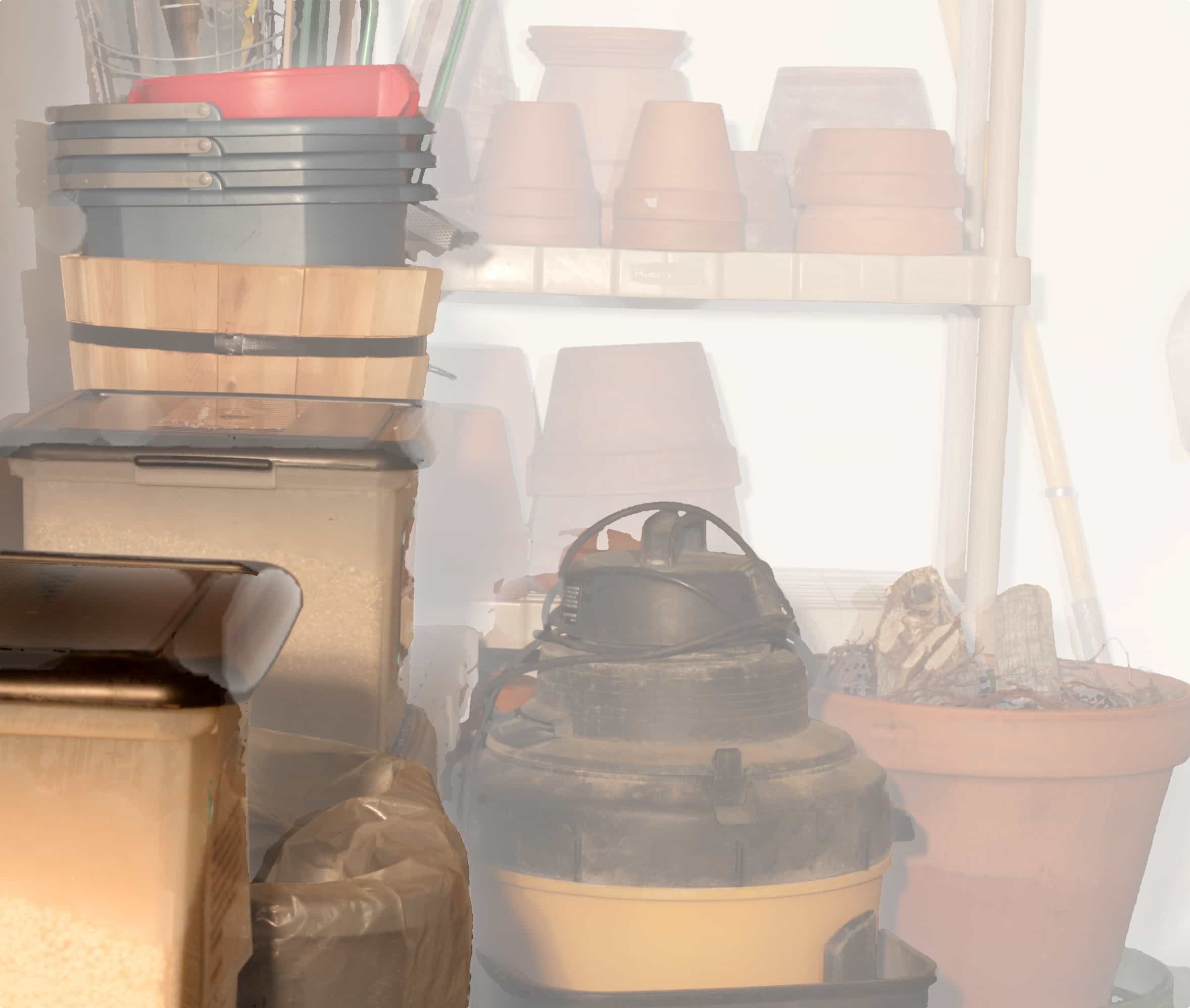}
        \caption{DehazeNet}
    \end{subfigure}
    \begin{subfigure}[t]{0.13\linewidth}
        \includegraphics[height=0.6in, width=\linewidth]{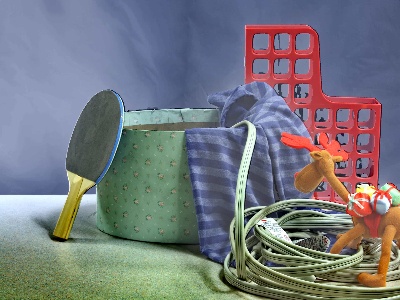} \\ 
        \includegraphics[height=0.6in, width=\linewidth]{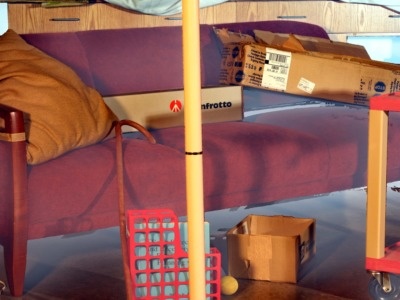} \\
        \includegraphics[height=0.6in, width=\linewidth]{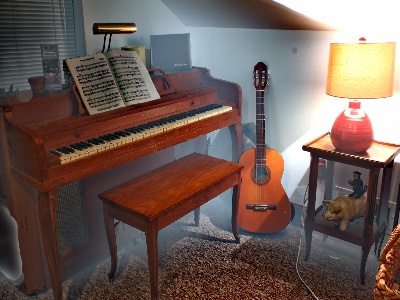} \\ 
        \includegraphics[height=0.6in, width=\linewidth]{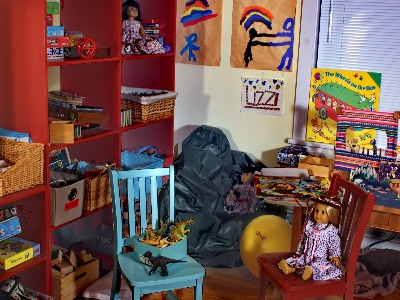} \\
        \includegraphics[height=0.6in, width=\linewidth]{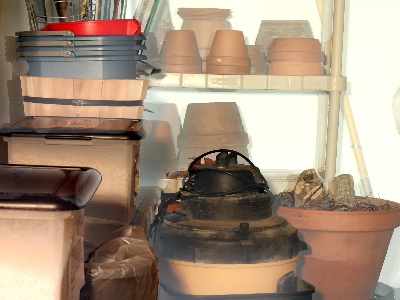}
        \caption{Berman \etal{}}
    \end{subfigure}
    \begin{subfigure}[t]{0.13\linewidth}
        \includegraphics[height=0.6in, width=\linewidth]{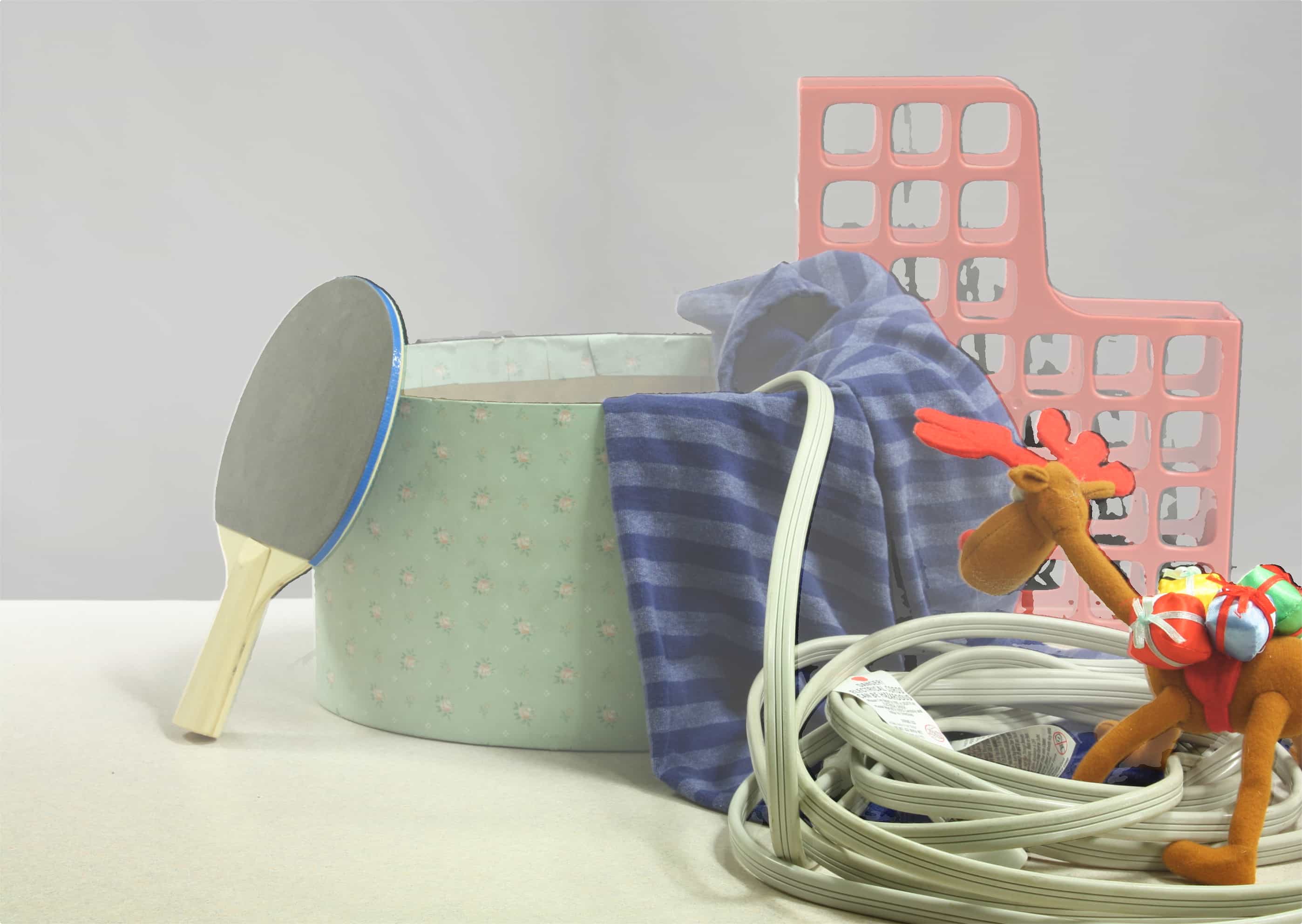} \\
        \includegraphics[height=0.6in, width=\linewidth]{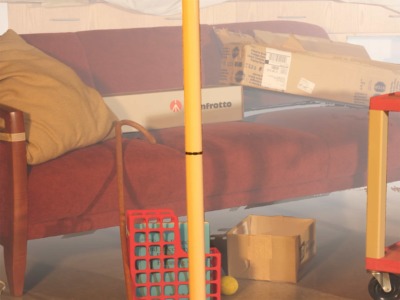} \\
        \includegraphics[height=0.6in, width=\linewidth]{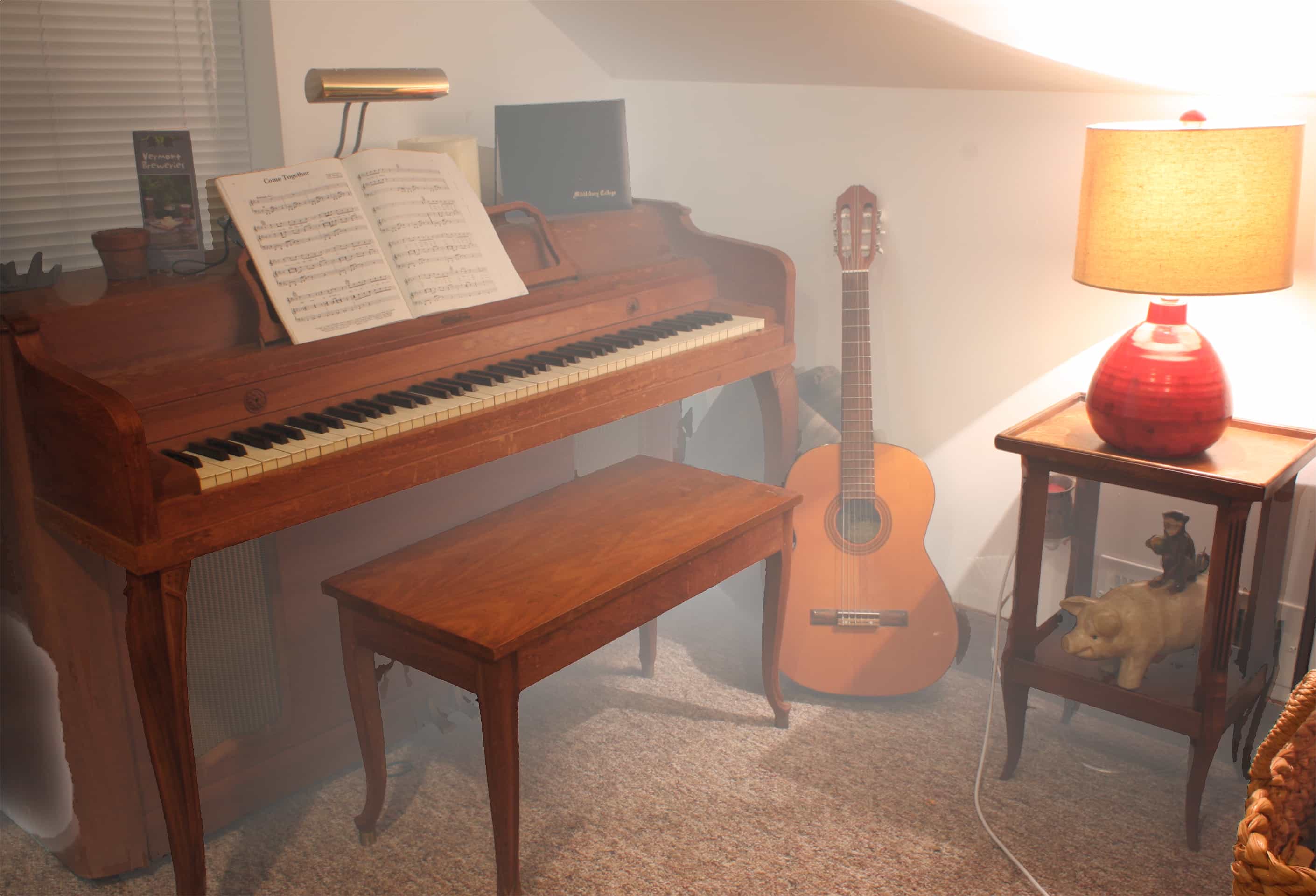} \\
        \includegraphics[height=0.6in, width=\linewidth]{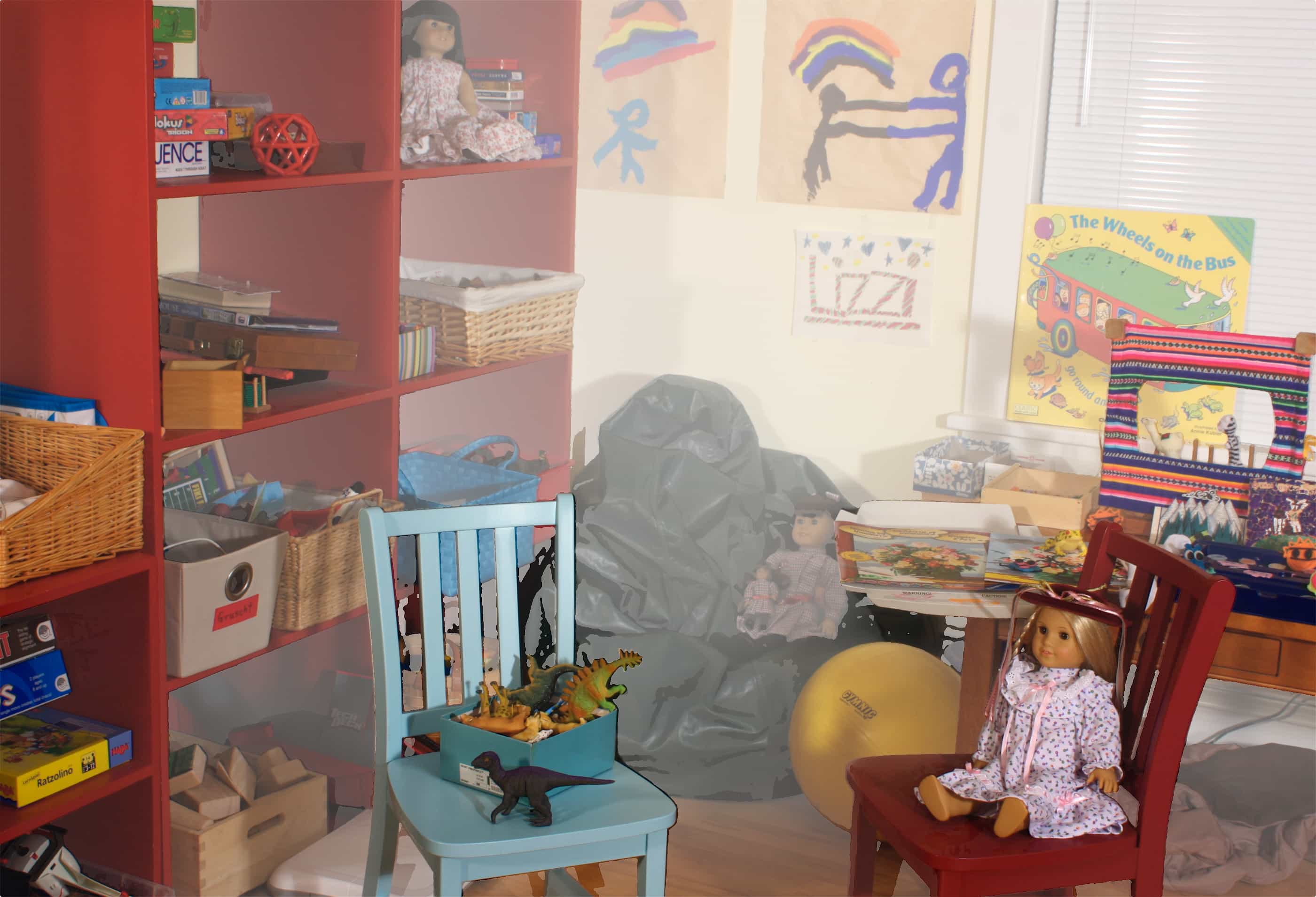} \\
        \includegraphics[height=0.6in, width=\linewidth]{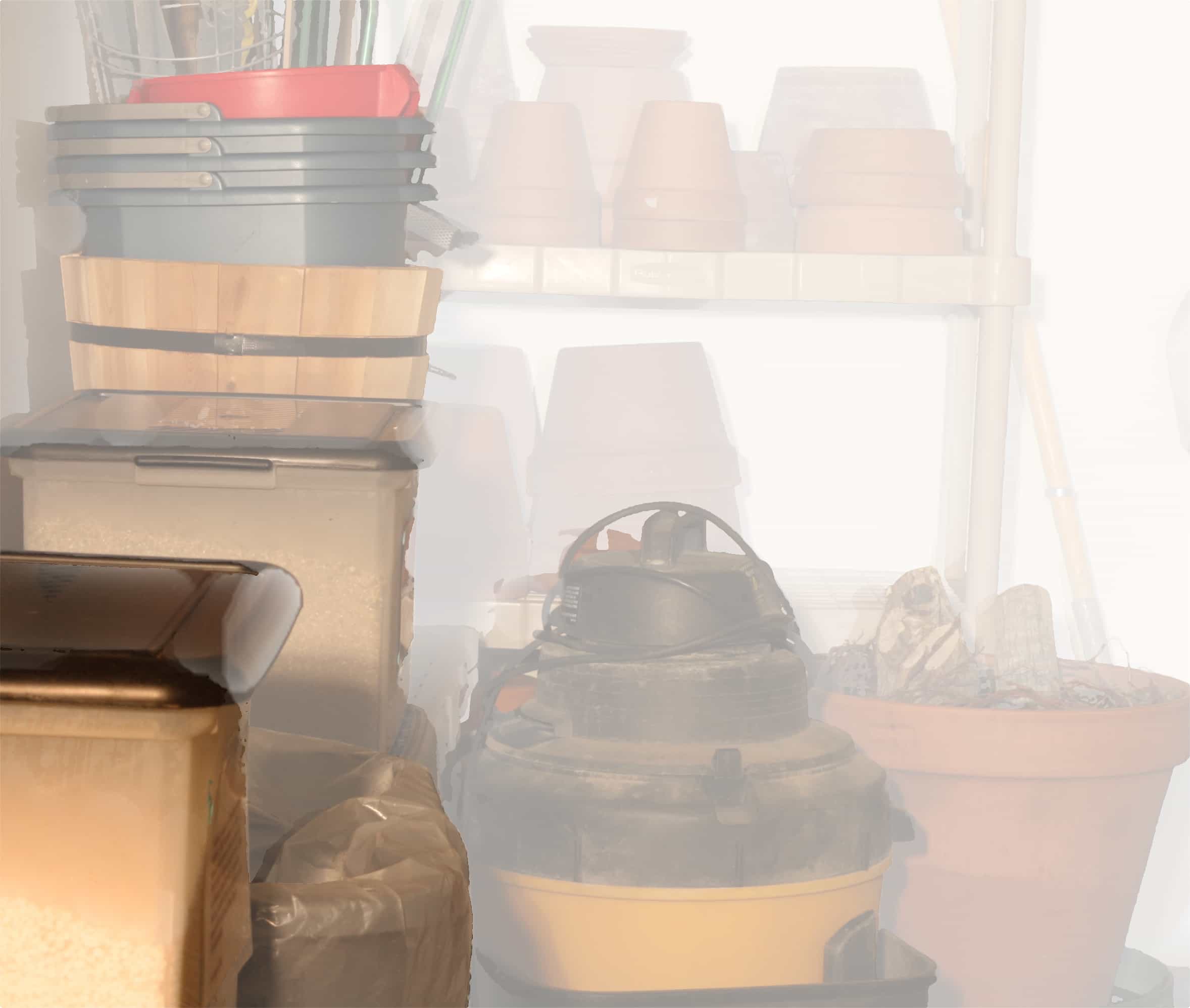}
        \caption{AOD-Net}
    \end{subfigure}
    \begin{subfigure}[t]{0.13\linewidth}
        \includegraphics[height=0.6in, width=\linewidth]{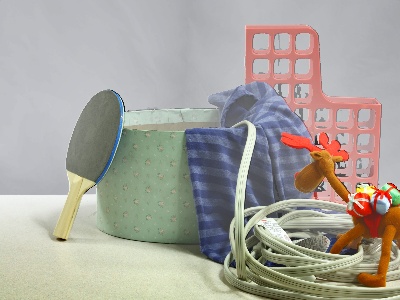} \\
        \includegraphics[height=0.6in, width=\linewidth]{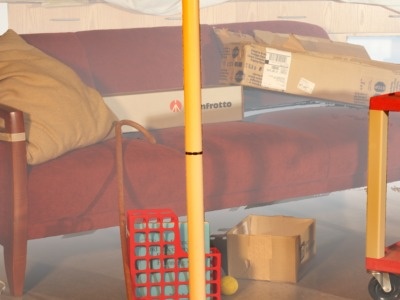} \\
        \includegraphics[height=0.6in, width=\linewidth]{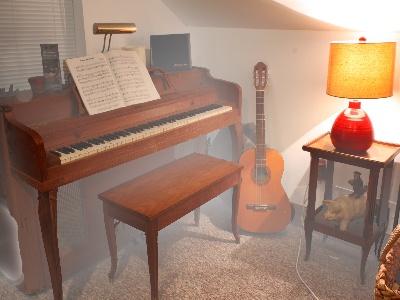} \\
        \includegraphics[height=0.6in, width=\linewidth]{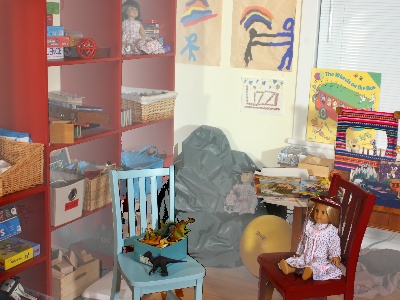} \\
        \includegraphics[height=0.6in, width=\linewidth]{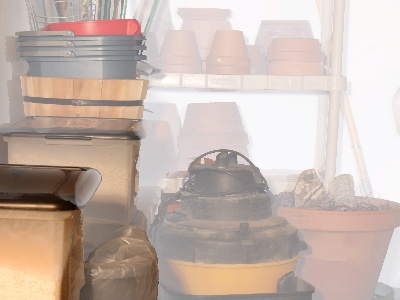}
        \caption{MSCNN}
    \end{subfigure}
    \begin{subfigure}[t]{0.13\linewidth}
        \includegraphics[height=0.6in,width=\linewidth]{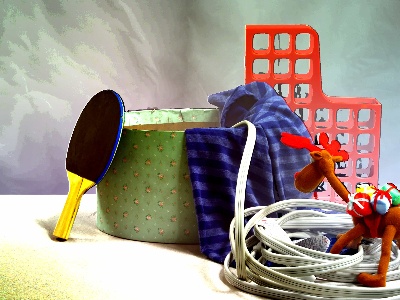} \\
        \includegraphics[height=0.6in,width=\linewidth]{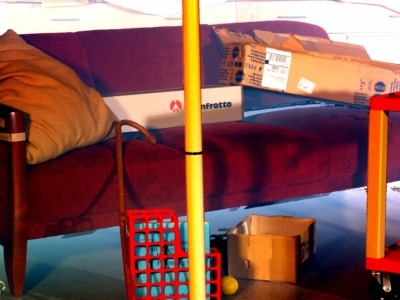} \\
        \includegraphics[height=0.6in,width=\linewidth]{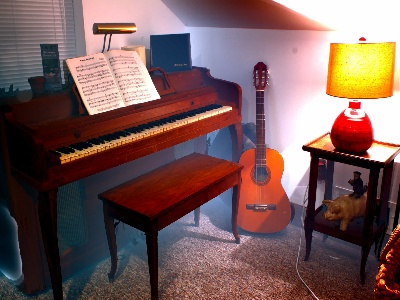} \\
        \includegraphics[height=0.6in,width=\linewidth]{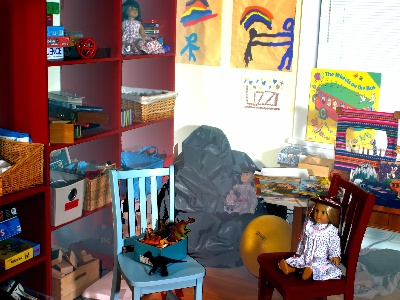} \\
        \includegraphics[height=0.6in,width=\linewidth]{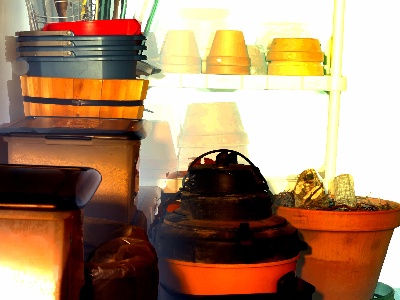}
        \caption{Ours}
    \end{subfigure}
    \begin{subfigure}[t]{0.13\linewidth}
        \includegraphics[height=0.6in,width=\linewidth]{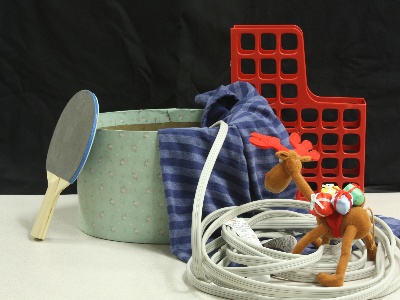} \\
        \includegraphics[height=0.6in,width=\linewidth]{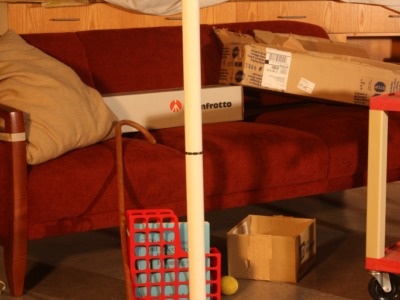} \\
        \includegraphics[height=0.6in,width=\linewidth]{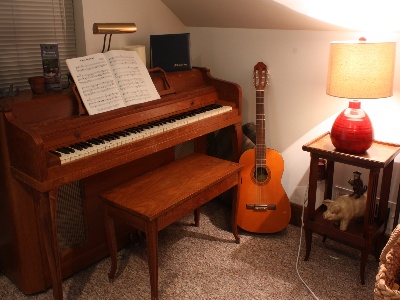} \\
        \includegraphics[height=0.6in,width=\linewidth]{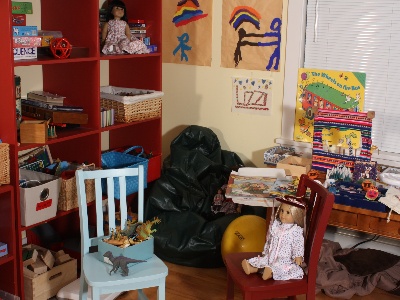} \\
        \includegraphics[height=0.6in,width=\linewidth]{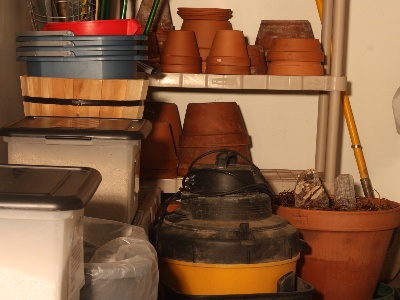}
        \caption{Ground truth}
    \end{subfigure}
    \caption{Visual comparison of images from Middlebury dataset \cite{middlebury}: \emph{Cable, Couch, Piano, Playroom, Shopvac}}
    \label{fig:middleburyvisual}
\end{figure*}

To measure how good an image has been dehazed we have used metrics like peak-signal-to-noise ratio (PSNR) and the structural similarity index (SSIM). A high value of these two indicates a better dehazed result. Apart from these two, we have used CIEDE2000 \cite{ciede} for measuring how well the colors have been restored. Its low value indicates that the resultant colors are close to the actual ones.


We demonstrate our results on Fattal's dataset in Figure. \ref{fig:fattalvisual}. We can observe from this figure that Dehazenet \cite{dehazenet} has not been able to properly dehaze the images, especially in cases of outdoor images (see \emph{Church, Lawn1 and Road1}). Berman \etal \cite{berman} is able to eliminate the haze to a certain extent, but it fails at removing dense haze (notice the background of \emph{Lawn1} and \emph{Road1}). AOD-Net \cite{aodnet} tends to saturate the images. Ren \etal \cite{ren} performs a little better but it retains more haze compared to Berman \etal{} \cite{berman}(see \emph{Lawn1 and Road1}). Our method has not only been able to remove the haze efficiently from both the foreground and the background, but it also does not hallucinate any colours. This is mainly because we estimate have estimated the $\mathbf{A}(\xcoord)$ correctly. The competence of our method is visible from the visual comparison and is also clearly indicated by the quantitative results in Table. \ref{tb:fattalquantitative}. 


For the Middlebury dataset, the comparisons are demonstrated in Fig. \ref{fig:middleburyvisual}. We notice that both MSCNN \cite{ren} and AOD-Net \cite{aodnet} have not been able to extenuate the haze fully and haze is visible specially at sharp edge discontinuities where they leave haze to a significant extent. Dehazenet \cite{dehazenet} removes haze a little better, but can not nullify it completely. The method of Berman \etal{} \cite{berman} and our method are successful in alleviating the haze from the images. Our method performs better when it comes to removing haze as visible from \emph{Cable, Couch, Piano and Shopvac} compared to Berman \etal{} \cite{berman}, but tends to saturate the images. Our method performs better than all the other CNN-based methods as validated by Table. \ref{tb:middquantitative}. But the over-saturation accounts for the lower Average PSNR and SSIM values when compared to Berman \etal{} \cite{berman}. 


\begin{figure*}
    \centering
    \begin{subfigure}[t]{0.15\linewidth}
        \includegraphics[width=\linewidth]{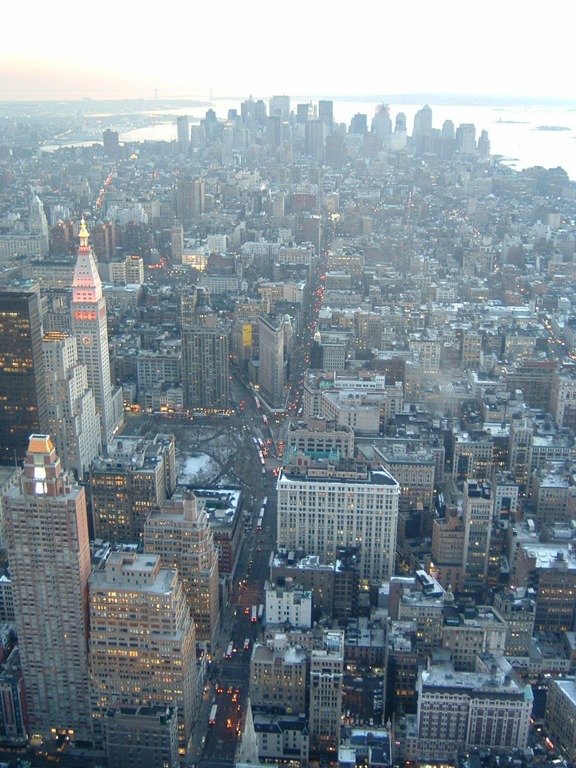}\\
        \includegraphics[width=\linewidth]{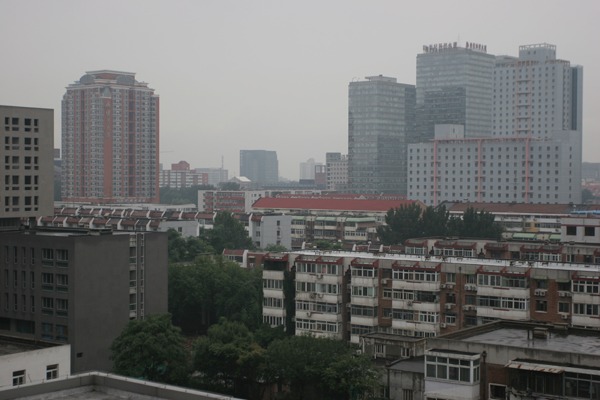}\\
        \includegraphics[width=\linewidth]{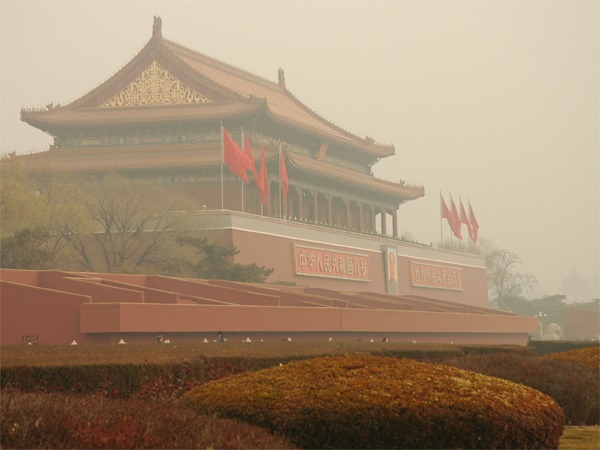}
        \caption{Hazy}
    \end{subfigure}
    \begin{subfigure}[t]{0.15\linewidth}
        \includegraphics[width=\linewidth]{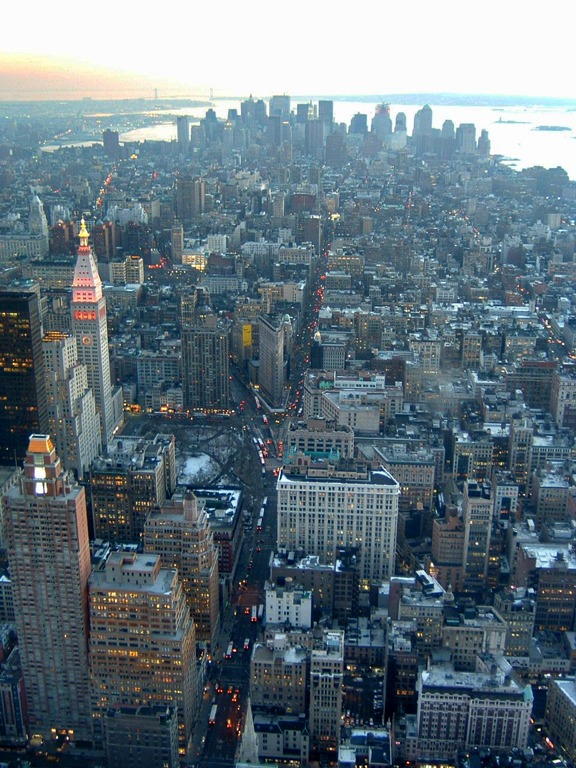}\\
        \includegraphics[width=\linewidth]{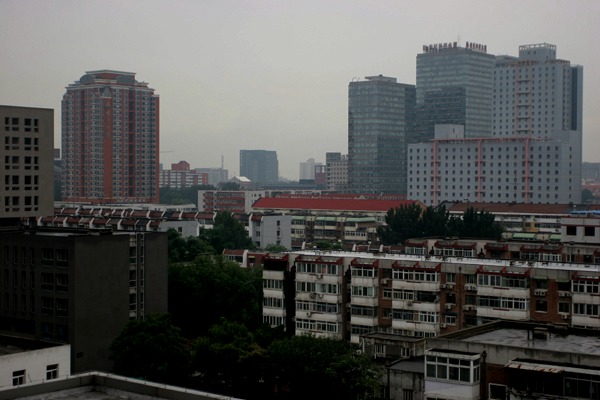}\\
        \includegraphics[width=\linewidth]{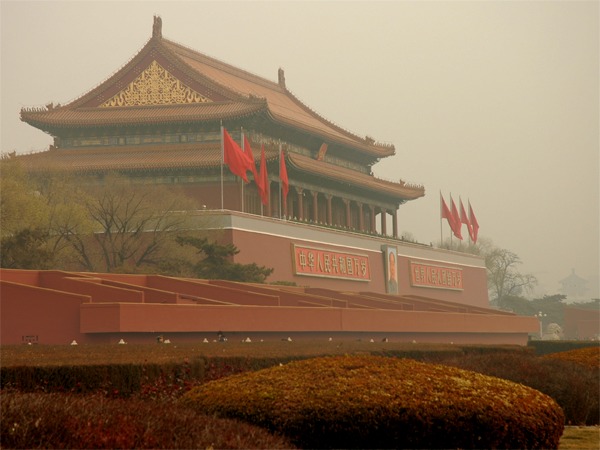}
        \caption{Dehazenet}
    \end{subfigure}
    \begin{subfigure}[t]{0.15\linewidth}
        \includegraphics[width=\linewidth]{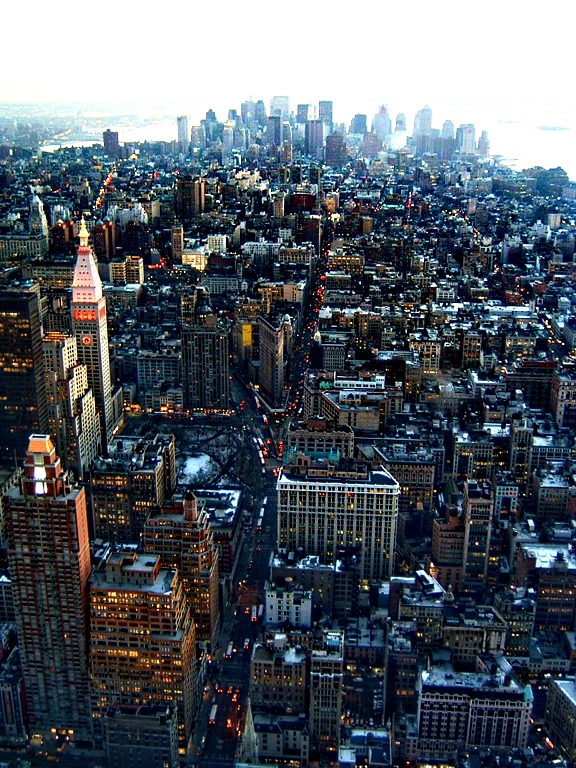}\\
        \includegraphics[width=\linewidth]{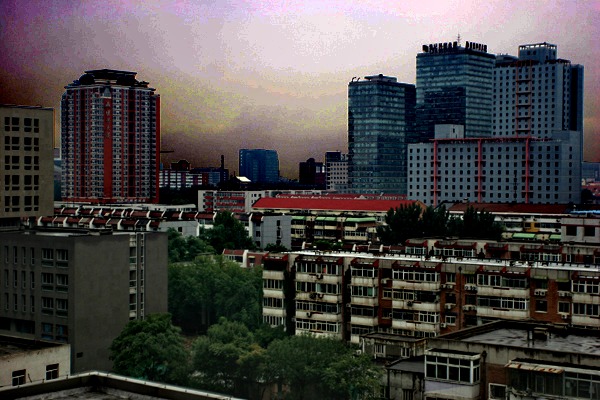}\\
        \includegraphics[width=\linewidth]{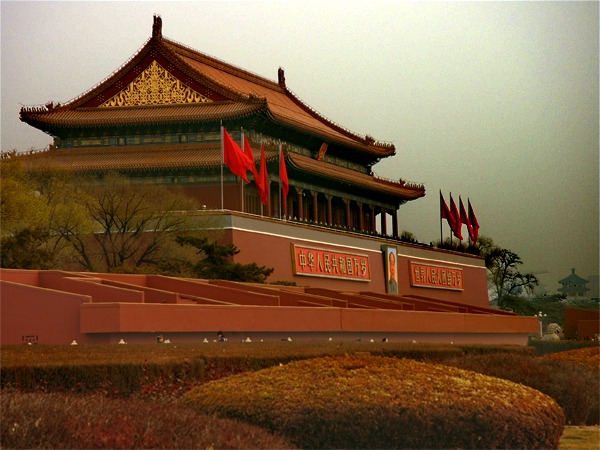}
        \caption{Berman \etal}
    \end{subfigure}
    \begin{subfigure}[t]{0.15\linewidth}
        \includegraphics[width=\linewidth]{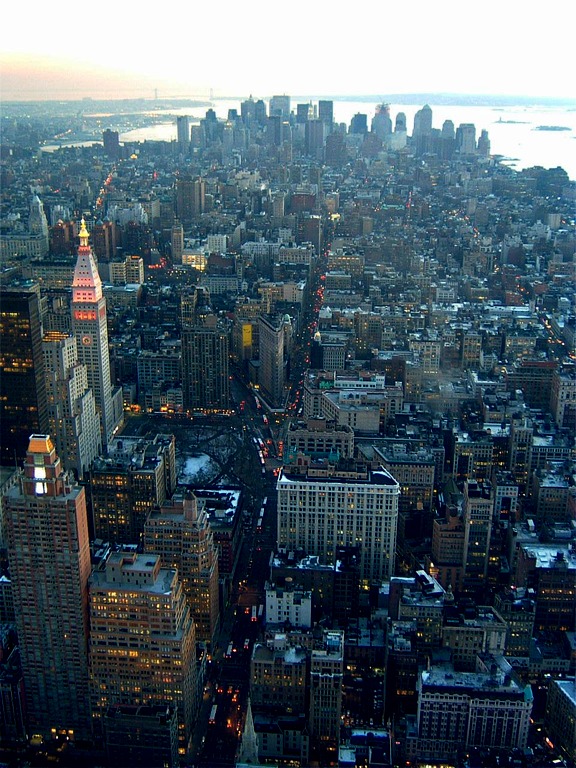}\\
        \includegraphics[width=\linewidth]{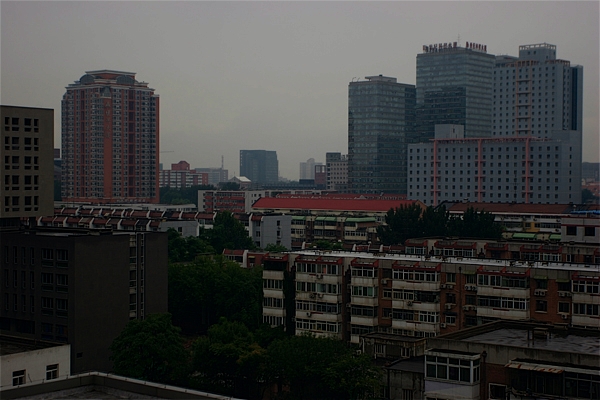}\\
        \includegraphics[width=\linewidth]{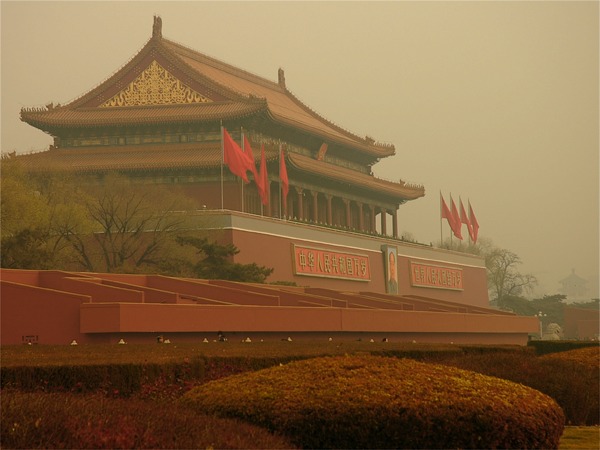}
        \caption{AOD-Net}
    \end{subfigure}
    \begin{subfigure}[t]{0.15\linewidth}
        \includegraphics[width=\linewidth]{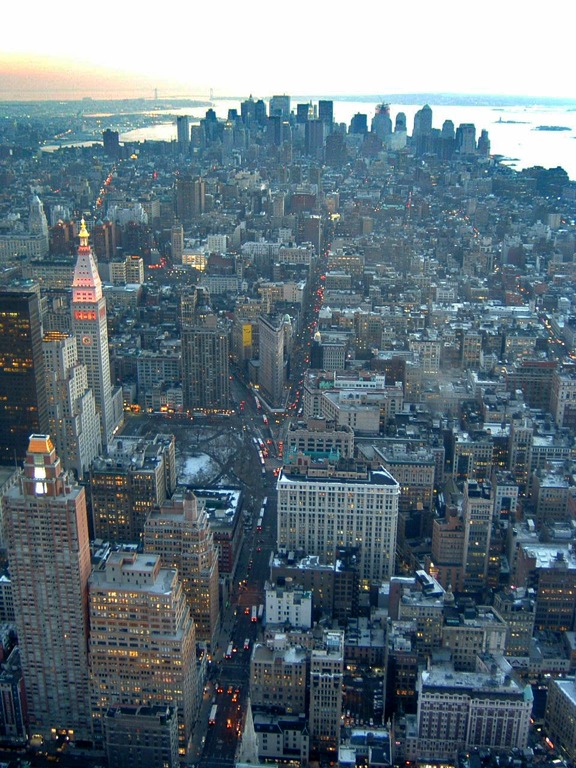}\\
        \includegraphics[width=\linewidth]{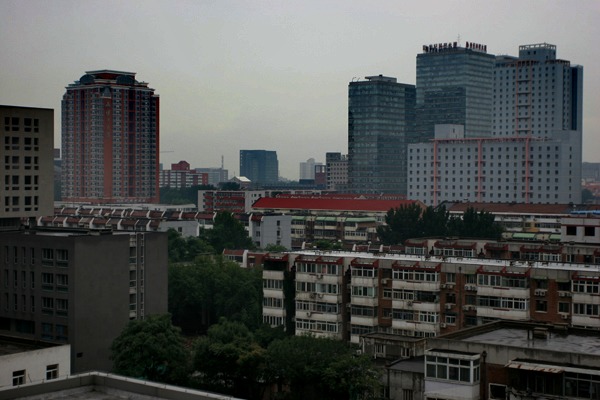}\\
        \includegraphics[width=\linewidth]{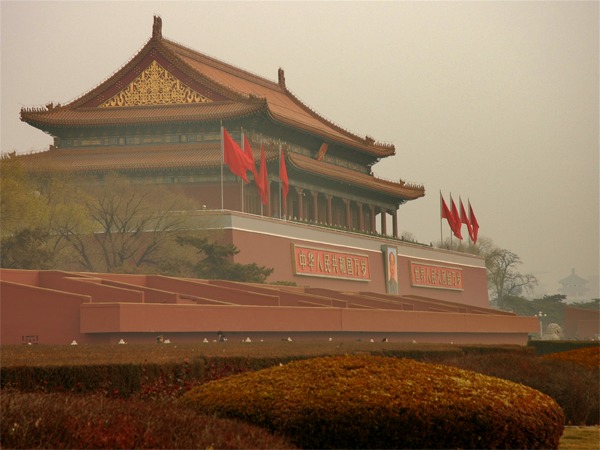}
        \caption{MSCNN}
    \end{subfigure}
    \begin{subfigure}[t]{0.15\linewidth}
        \includegraphics[width=\linewidth]{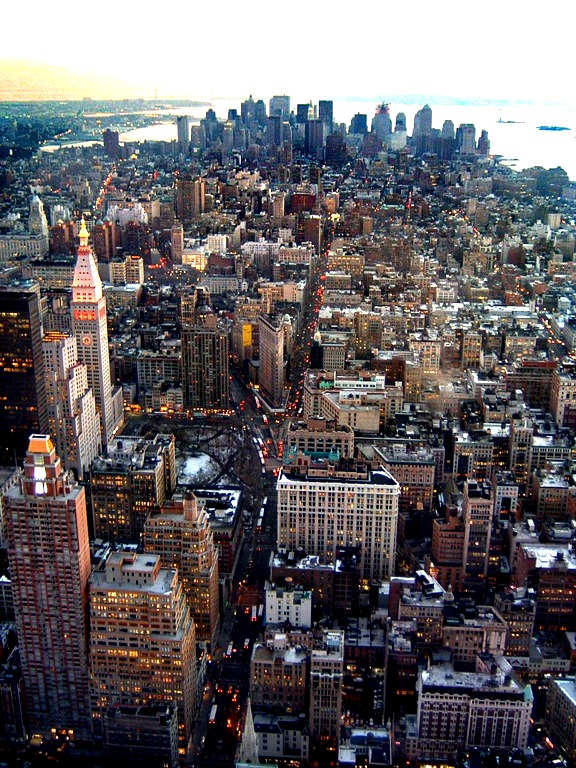}\\
        \includegraphics[width=\linewidth]{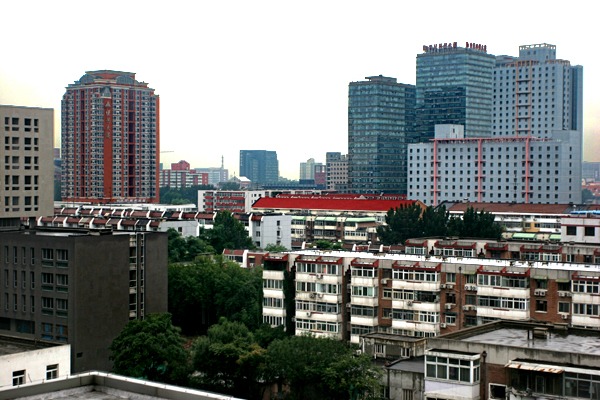}\\
        \includegraphics[width=\linewidth]{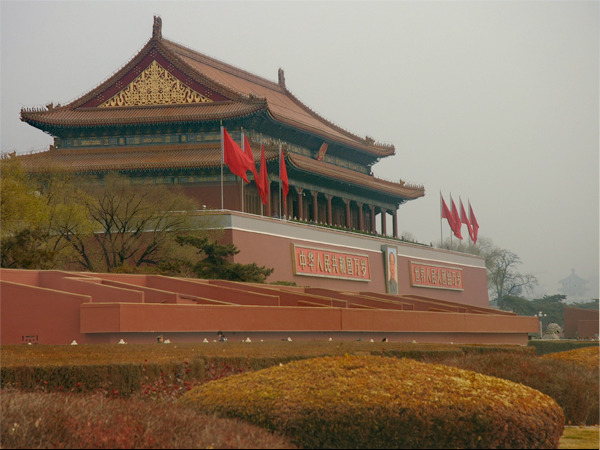}
        \caption{Ours}
    \end{subfigure}
    
        %
    \caption{Visual comparison on real-world images: \emph{New York, Building} and \emph{Tiananmen}}
    \label{fig:realworld}
\end{figure*}

\begin{figure}
    \centering
    \begin{subfigure}[t]{0.24\linewidth}
        \includegraphics[width=\linewidth]{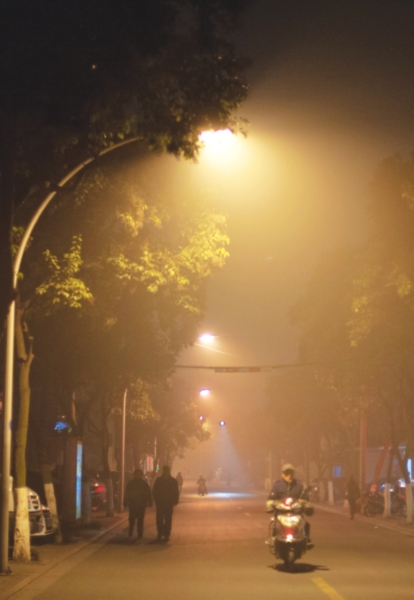}\\
        \includegraphics[width=\linewidth]{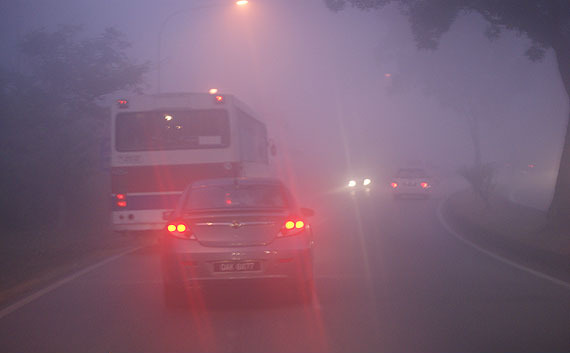}
        \caption{Haze}
    \end{subfigure}
    \begin{subfigure}[t]{0.24\linewidth}
        \includegraphics[width=\linewidth]{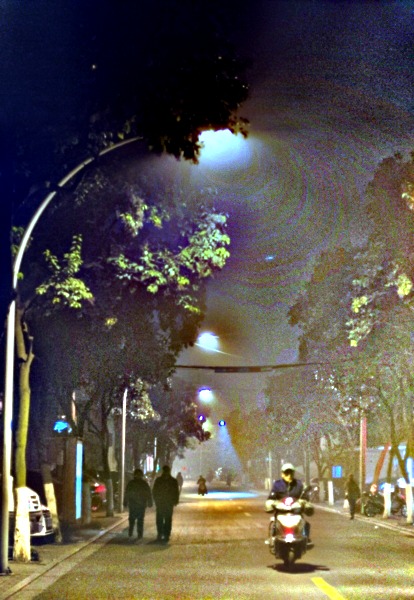}\\
        \includegraphics[width=\linewidth]{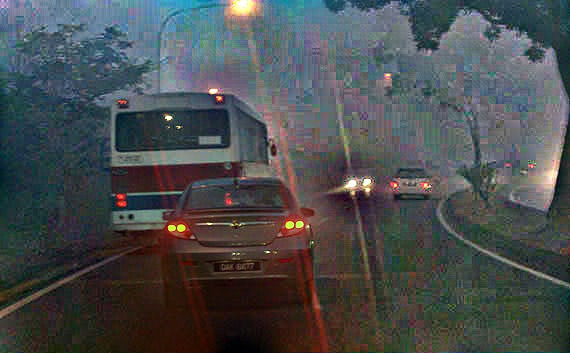}
        \caption{Li \etal{}}
    \end{subfigure}
    \begin{subfigure}[t]{0.24\linewidth}
        \includegraphics[width=\linewidth]{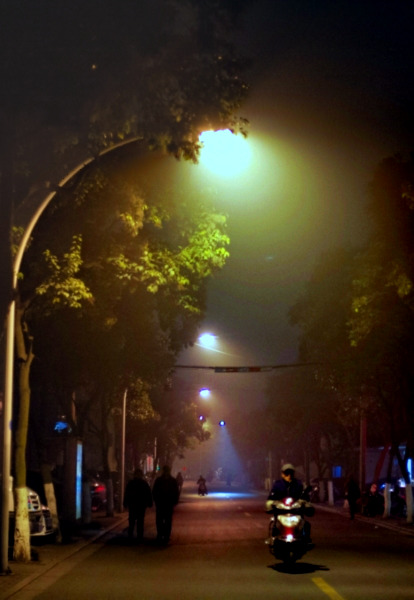}\\
        \includegraphics[width=\linewidth]{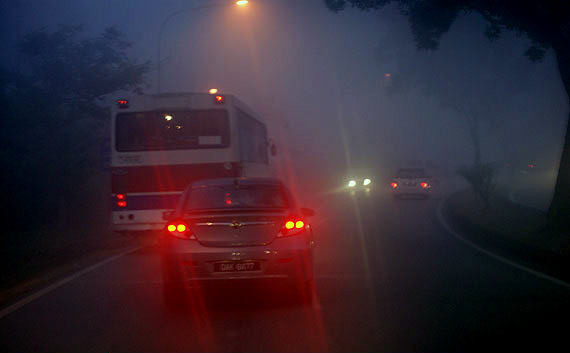}
        \caption{Santra and Chanda}
    \end{subfigure}
    \begin{subfigure}[t]{0.24\linewidth}
        \includegraphics[width=\linewidth]{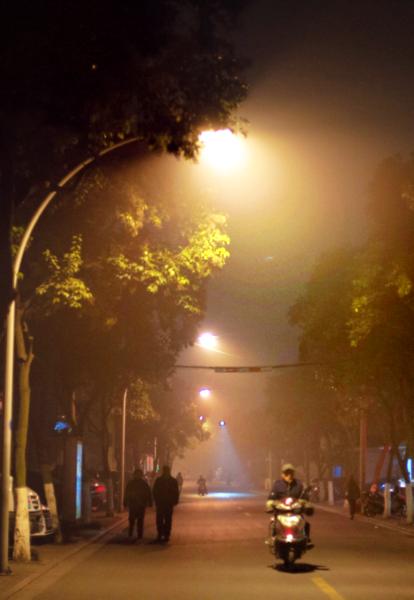}\\
        \includegraphics[width=\linewidth]{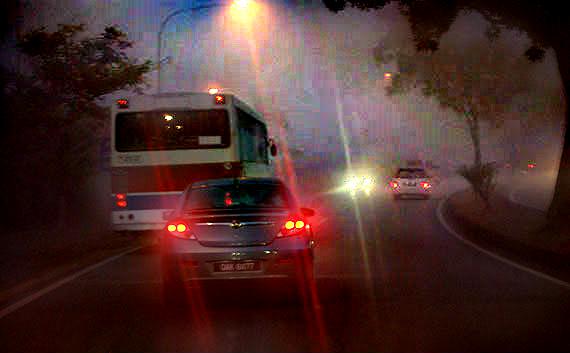}
        \caption{Ours}
    \end{subfigure}
    \caption{Comparison of night-time dehazing}
    \label{fig:night}
\end{figure}

\subsection{Real World Images}
\label{sub:real}
In order to establish the efficacy of our model, we have qualitatively compared the results of dehazing real-world benchmark images. We have also include a comparison of night-time dehazing. Figure.\ref{fig:realworld} shows the result on real-world hazy images: \emph{New York, Building} and \emph{Tiananmen}. Dehazenet \cite{dehazenet} and MSCNN \cite{ren} are not able to clear the haze layer entirely due to under-estimation of the thickness of the haze. Due to this, dehazing results from both the methods tend to have a dull contrast (specifically visible in \emph{New York} and \emph{Tiananmen}). Berman \etal \cite{berman} is able to mitigate the haze effectively and enhance the visibility. However, in the process, it over-saturates the contrast and tends to produce some colour distortions. For example see the sky region of \emph{New York} which appears to be whiter than it actually is and occludes the top of the skyscrapers. Also in \emph{Building}, \cite{berman} tends to hallucinate a light-purple colour for the skies. The results from AOD-Net \cite{aodnet} do not produce any colour distortions or unwanted artifacts in the first two images but leaves some haze. While in \emph{Tiananmen}, it envelopes the whole image by a yellowish layer. In distinction, our method produces images which are comparatively the least hazy while maintaining the clarity, colour and contrast composition and keeping the crisp details intact.



Our method has been designed taking into consideration the scenario of night-time dehazing. To establish the effectiveness of our method in this situation we provide qualitative comparison with night-time dehazing methods \cite{li_night, santra2016day}. Fig. \ref{fig:night} exhibits some comparison . Li \etal \cite{li_night} tends to over-sharpen the images and create noise in form of grains that is visible especially around the areas of illumination. The dehazed images from Santra and Chanda \cite{santra2016day} deviates from the normal colour composition by over-saturation and anomalous colour hallucination as visible from the results. The light from the street light is yellow but the result displays a green tinge. Our proposed method is efficient in the removal of haze without introducing of artifacts and colour incoherence. 


\begin{table*}
\caption{Average PSNR/SSIM/CIEDE2000 values of different loss function on Fattal's and Middlebury dataset}
\label{tab:ablation_quan}
\begin{tabularx}{\textwidth}{X|X X X X X X}
\hline
Dataset & MSE  & $L_3$  & $L_1$+$L_2$   & $L_2$+$L_3$  & $L_1$+$L_3$  & $L_1$+$L_2$+$L_3$ \\\hline
Fattal           & 15.5/0.5/19.4 & 12.6/0.3/20.5 & 16.1/0.5/17.0 & 6.3/0.2/34.8  & 12.8/0.4/22.0 & \textbf{18.2/0.8/15.8} \\
Middlebury       & 8.6/0.4/32.0  & 6.6/0.3/38.5  & 8.0/0.4/33.0  & 3.4/0.3/53.1 & 7.35/0.4/35.6   & \textbf{11.8/0.6/21.7} \\\hline
\end{tabularx}
\end{table*}

\begin{figure*}
    \centering
    \begin{subfigure}[t]{0.13\linewidth}
        \includegraphics[width=\linewidth]{images/fattal/hazy/lawn1_input.jpg}\\
        \includegraphics[width=\linewidth]{images/middlebury/hazy/Piano_Hazy.jpg}
        \caption{Hazy}
    \end{subfigure}
    \begin{subfigure}[t]{0.13\linewidth}
        \includegraphics[width=\linewidth]{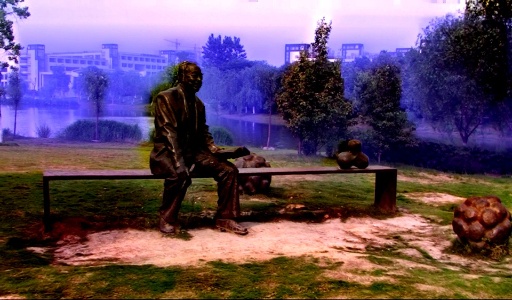}\\
        \includegraphics[width=\linewidth]{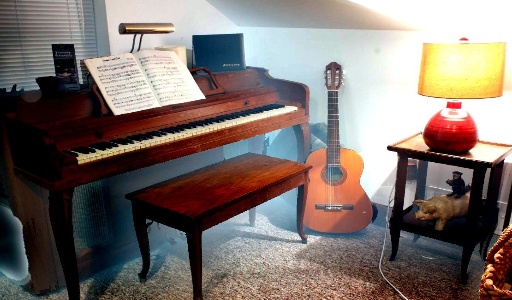}
        \caption{MSE}
    \end{subfigure}
    \begin{subfigure}[t]{0.13\linewidth}
        \includegraphics[width=\linewidth]{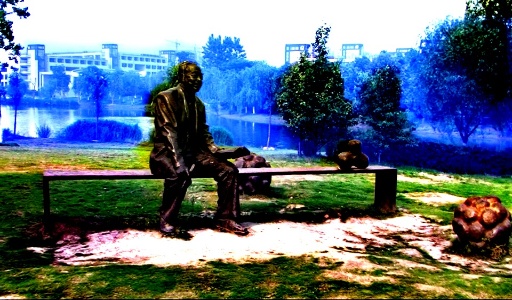}\\
        \includegraphics[width=\linewidth]{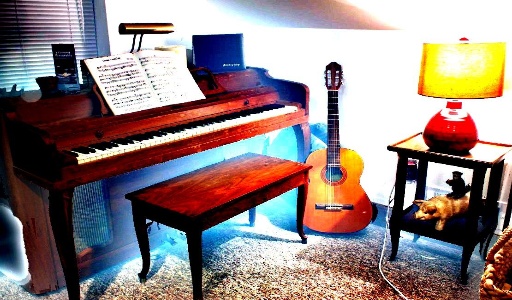}
        \caption{$L_3$}
    \end{subfigure}
    \begin{subfigure}[t]{0.13\linewidth}
        \includegraphics[width=\linewidth]{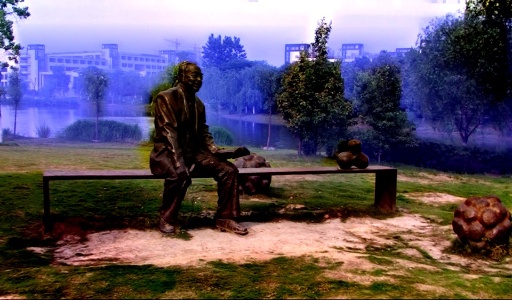}\\
        \includegraphics[width=0.97\linewidth]{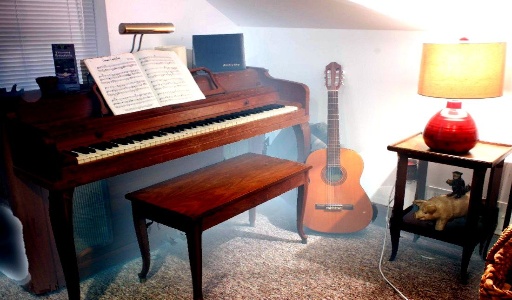} 
        \caption{$L_1$+$L_2$}
    \end{subfigure}
    \begin{subfigure}[t]{0.13\linewidth}
        \includegraphics[width=\linewidth]{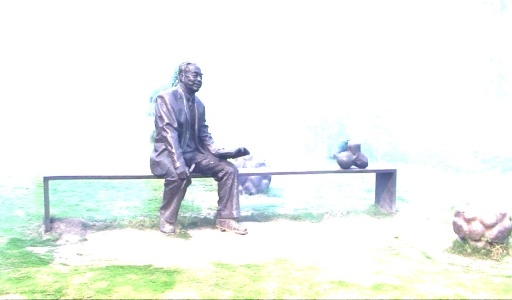}\\
        \includegraphics[width=0.97\linewidth]{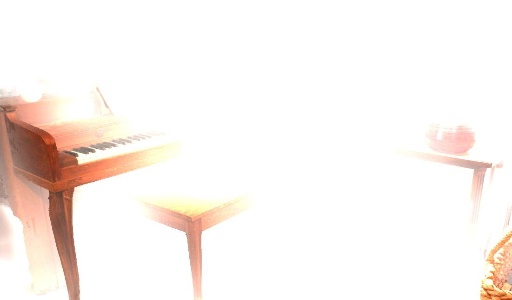} 
        \caption{$L_2$+$L_3$}
    \end{subfigure}
    \begin{subfigure}[t]{0.13\linewidth}
        \includegraphics[width=\linewidth]{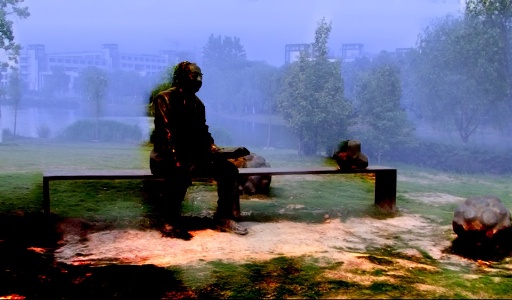}\\
        \includegraphics[width=\linewidth]{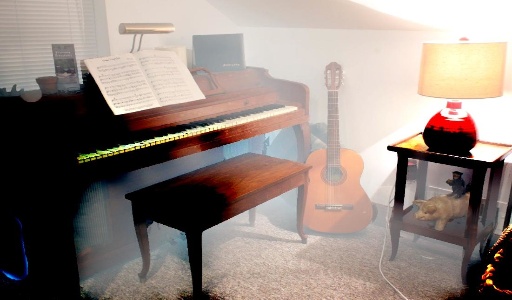} 
        \caption{$L_1$+$L_3$}
    \end{subfigure}
    \begin{subfigure}[t]{0.13\linewidth}
        \includegraphics[width=\linewidth]{images/fattal/ours/lawn1_input.jpg}\\
        \includegraphics[width=\linewidth]{images/middlebury/ours/Piano_Hazy.jpg}
        \caption{$L_1$+$L_2$+$L_3$}
    \end{subfigure}
    \caption{Visual comparison of \textit{Lawn2} from Fattal's dataset and \textit{Piano} from Middlebury dataset using different loss functions}
    \label{fig:ablation_qual}
\end{figure*}

\subsection{Failure Cases}
\label{sub:failure}

\begin{figure}
    \centering
    \includegraphics[width=0.23\linewidth]{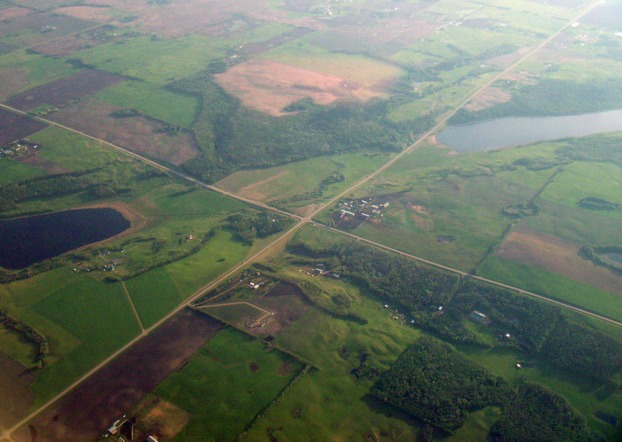} \includegraphics[width=0.23\linewidth]{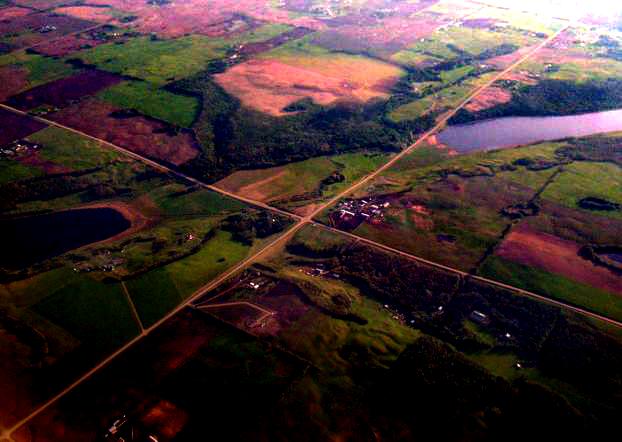} \includegraphics[width=0.23\linewidth]{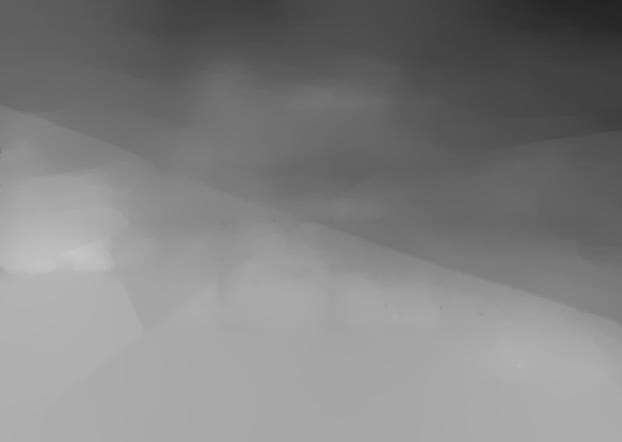}
    \includegraphics[width=0.23\linewidth]{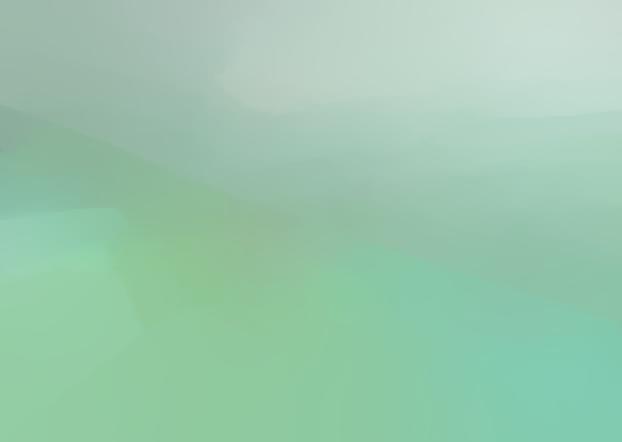}
    \includegraphics[width=0.23\linewidth]{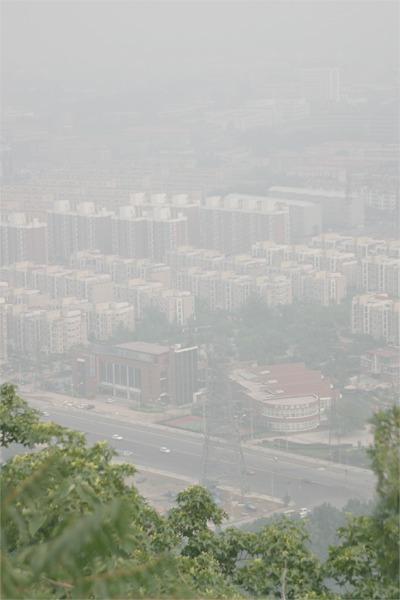} \includegraphics[width=0.23\linewidth]{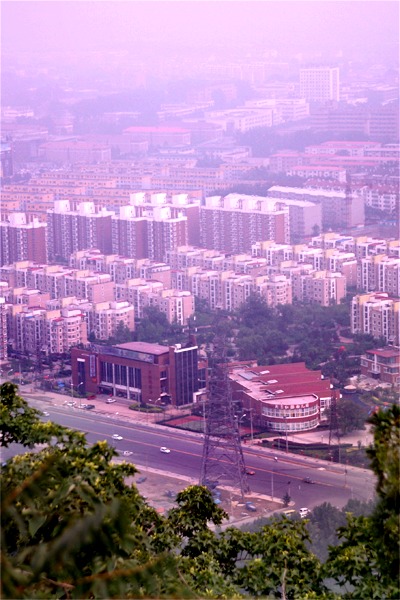} \includegraphics[width=0.23\linewidth]{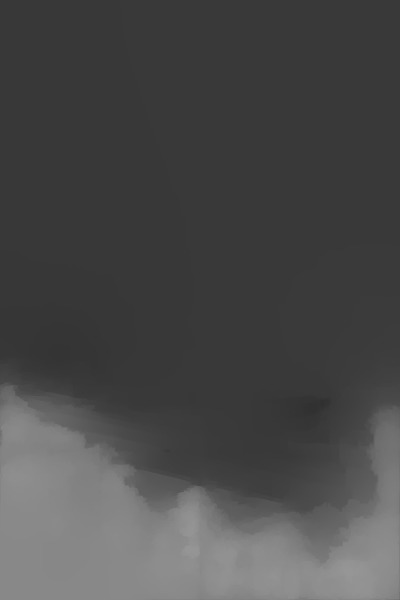}  
    \includegraphics[width=0.23\linewidth]{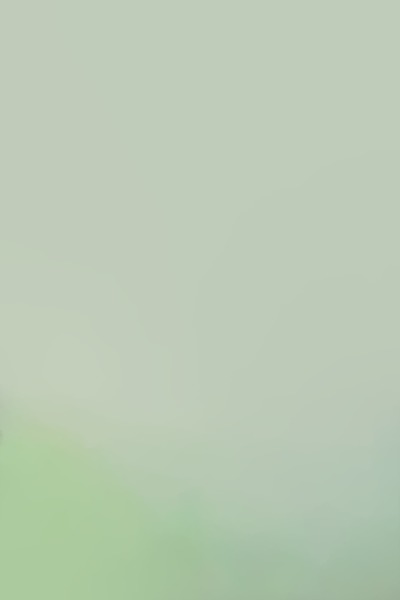}
    \caption{Failure on \emph{aerial}(top) and \emph{cityscape}(bottom). \textit{Left to Right:} Input image, dehazed image, scene transmittance map, airlight map}
    \label{fig:fig_fail}
\end{figure}
Our proposed method performs well in diverse lighting conditions and both indoor and outdoor situations which is evident from our evaluations. However, there are some cases in which our model fails to produce satisfactory results. We demonstrate some examples in Figure. \ref{fig:fig_fail}. In \emph{aerial} image, our method is able to estimate the scene transmittance correctly but changes the colour of the final dehazed image due to incorrect \airlight estimation. While in \emph{cityscape}, our method is not able to estimate the scene transmittance correctly. As we can observe that the scene transmittance tends to stay constant after a certain point which is why the removal of haze is inadequate in the distant parts of the image. The \airlight map is also anomalous which accounts for the purple shade of the dehazed output. 

\subsection{Ablation Studies}
\label{sub:ablation}
We have already stated, we use a custom defined loss function to train our network. In this subsection, we quantitatively show the improvement we get from moving away from \emph{MSE}. We also show all the three components of our loss is necessary for correct estimation. To compare, we train our network with each of the following losses independently in addition to our original loss ($L_1$, $L_2$ and $L_3$): \emph{MSE} loss on $\mathbf{A}$ and $t$, only $L_3$, $L_1$ and $L_2$, $L_2$ and $L_3$, $L_1$ and $L_3$. Trained with each of the losses, we compute the PSNR, SSIM and CIEDE2000 values for both Fattal and Middlebury dataset. Table. \ref{tab:ablation_quan} show the quantitative results. For visual comparison, we have included result of two images in Fig. \ref{fig:ablation_qual}. In can be seen, the model trained using $L_2$ and $L_3$ tends to estimate very small values for the transmittance and as a result the output is almost white. This is because the network was trained without the supervision of ground-truth $t(\xcoord)$. For \emph{MSE} loss the images are dehazed where the haze is thin, but it had failed where haze is thick. This validates our observation that MSE can fail at places where the value of $t(\xcoord)$ is small (thick haze). The model with $L_3$ only achieves similar results to \emph{MSE}, but the color is worse as it has not been able to estimate the environmental illumination correctly. \emph{MSE} performs better in this regard as it had access the ground-truth illumination during training. Networks trained on  $L_1$ + $L_3$ produce very bad results especially in outdoor images as it can be seen in \emph{Lawn 2}. This happens because the network  was trained without the supervision of ground-truth $\mathbf{A}(\xcoord)$ as a result it could not estimate the environmental illumination. This also shows the dependence of the two haze parameters. The network trained with $L_1$ + $L_2$ gives the second best result as the network learns with ground-truth $t(\xcoord)$ and $\mathbf{A}(\xcoord)$. But it could not match the combination of all three losses as it didn't capture the dependence of the two parameters.




\section{Conclusions}
\label{sec:conclusion}
In this paper, we have proposed to approach the problem of image dehazing by jointly estimating the scene transmittance and the environmental illumination map from image patches. Haze-relevant features are extracted using a two-way forked FCN that is trained by minimizing a novel loss function. The loss is based on the imaging model, therefore, it takes into consideration the relationship between scene transmittance and the environmental illumination. We have also shown that all parts of the loss function is necessary for correct estimation of the haze parameters. Although the method estimates the environmental illumination, due to patch based processing it fails at some cases. Using full images to estimate the environmental illumination can improve the results. While the FCNs can work independent of input image size but the trained network always depends on the scale of training data. This issue is yet to be solved.

\bibliographystyle{plainnat}
\bibliography{template}

\end{document}